\documentclass{article}

\usepackage{microtype}
\usepackage{graphicx}
\usepackage{tabularx}
\usepackage{threeparttablex}
\usepackage{subcaption}
\usepackage{booktabs} % for professional tables
\usepackage{enumitem}
\usepackage{tcolorbox}

\usepackage{hyperref}

\usepackage[accepted]{icml2025}

\usepackage{amsmath}
\usepackage{amssymb}
\usepackage{mathtools}
\usepackage{amsthm}

\usepackage[capitalize,noabbrev]{cleveref}

\theoremstyle{plain}

\theoremstyle{definition}

\theoremstyle{remark}

\usepackage[textsize=tiny]{todonotes}

\usepackage{xcolor}

\definecolor{deepblue}{RGB}{0,0,139} % or adjust to your desired shade

\icmltitlerunning{Best-of-Q: Improving VLM Agents with Q-function action ranking at inference}

\begin{document}

\twocolumn[
\icmltitle{Best-of-Q: Improving VLM Agents  \\
           with Q-function Action Ranking at Inference}



\icmlsetsymbol{intern}{\dag}
\icmlsetsymbol{equal}{*} % Keeping your existing equal contribution symbol

\begin{icmlauthorlist}
    % 2. Add the 'intern' tag to the author (Emilien)
    \icmlauthor{Emilien Biré}{hc} 
    \icmlauthor{María Santos}{hcp}
    \icmlauthor{Kai Yuan}{hc}
\end{icmlauthorlist}

\icmlaffiliation{hc}{H Company, Paris, France}
\icmlaffiliation{hcp}{Work done while at H Company, now unaffiliated}

% 3. Add the text for the internship symbol
\icmlcorrespondingauthor{Emilien Biré}{emilien.bire@student-cs.fr}
%\icmlcorrespondingauthor{María Santos}{maria@hcompany.ai}
%\icmlcorrespondingauthor{Kai Yuan}{kai.yuan@hcompany.ai}

\icmlkeywords{Machine Learning}

\vskip 0.3in
]



\printAffiliationsAndNotice{} % otherwise use the standard text.

\begin{abstract}
Vision-Language Models (VLMs) have become powerful backbones for agents to autonomously operate in digital environments like the web and operating systems.
However, these models suffer from inadaptability to fast-changing environments like the web, which can be alleviated by fine-tuning requiring expansive model training and data collection. In this work, we introduce a novel paradigm for enhancing agentic VLM policies at inference without policy retraining. Fundamentally, our approach decouples the VLM's role as a high-capacity action proposer from the final action selection mechanism. We keep the VLM policy frozen and use it to generate a set of candidate actions for a given state. Then, a lightweight, offline-trained Q-function reranks these candidates, and the agent executes the action with the highest estimated value. The main contribution is to apply the Q-function directly during inference for immediate policy improvement, and not offline to relabel data for policy retraining. We demonstrate on the academic WebVoyager benchmark that our method significantly boosts agent success rates, improving a Qwen2.5-VL-7B agent from 38.8\% to 55.7\% and a proprietary GPT-4.1 agent from 82.4\% to 88.8\%.
\end{abstract}

\section{Introduction}
\label{sec:introduction}

The quest to build autonomous agents that can intelligently operate complex digital interfaces, such as websites and computers, has seen remarkable progress with the advent of large foundation models. In particular, Vision-Language Models (VLMs) have emerged as powerful policy backbones for these agents, capable of parsing multimodal inputs (e.g., screenshots and text) and generating low-level actions to accomplish high-level user goals \cite{he_webvoyager_2024, hong_cogagent_2024}
% \textcolor{red}{Add more refs and different applications (device control, gaming, robots, ...}
. These models, often pre-trained on vast internet-scale data, possess a strong semantic understanding of visual elements and user intent, making them a natural fit for tasks requiring navigation and interaction within graphical user interfaces (GUIs). Frameworks like WebVoyager \cite{he_webvoyager_2024}, WebArena \cite{zhou_webarena_2024}, and OSWorld \cite{xie_osworld_2024}, AndroidWorld \cite{androidworld2025}, Windows Agent Arena \cite{windowsagentarena2024} have demonstrated the potential of VLM-based agents to tackle a diverse range of real-world computer and web tasks.

Despite being good initial policies, most VLMs are not geared towards action planning and achieving state-of-the-art performance on web browsing tasks often requires further adaptation through methods like Reinforcement Learning (RL) or large-scale Imitation Learning (IL). On-policy RL methods are constrained by their reliance on extensive and costly online interaction to gather sufficient data for stable policy gradient updates \cite{mobileguirl2025}. This requirement is often impractical in real-world settings where interaction, such as navigating a website, is inherently slow and expansive. Imitation learning from expert demonstrations offers an alternative that bypasses the need for online interaction. However, this approach introduces its own significant bottleneck: the manual and resource-intensive process of collecting a large, high-quality, and diverse dataset of expert trajectories. Furthermore, those methods imply the iterative fine-tuning of a billion-parameter VLM, and the computational expense of this step renders each training cycle a significant investment.

In this work, we propose an alternative paradigm for enhancing the performance of pre-trained VLM policies without the need for expensive retraining or online interaction. Our central idea is to separate the VLM's role in perception and action generation from the distinct process of strategic, value-based decision-making. This approach allows the agent to identify the optimal path forward while making fewer end-to-end reasoning demands on the potentially flawed and unspecialized VLM. Our method employs a \textit{Best-of-N} reranking mechanism at the point of action selection. For any given state (e.g., a screenshot of a webpage), we first use the VLM policy to generate a diverse set of $N$ candidate actions. Then, our trained Q-function evaluates each of these candidates and assigns them a Q-value, which represents the expected future return of taking that action. The agent executes the action with the highest predicted Q-value. This allows the agent to dynamically select more promising actions from the VLM's proposed set, correcting potential mistakes or suboptimal choices made by the base policy. This approach circumvents the high cost of VLM fine-tuning while still enabling substantial policy improvement at \textit{inference time}.

Our main contributions are:
\begin{enumerate}[leftmargin=*, nosep]
    \item A novel framework for improving pre-trained VLM agents that separates policy generation from value-based action selection, avoiding costly VLM fine-tuning.
    \item The application of a Best-of-N reranking mechanism at inference time, guided by an offline-trained Q-function, to enhance the decision-making of a VLM policy.
    \item An extensive empirical evaluation on the WebVoyager benchmark demonstrating that our method:
    \begin{itemize}[nosep]
        \item[a)] Significantly boosts success rates (e.g., from 38.8\% to 55.7\% for Qwen2.5-7B).
        \item[b)] Offers a superior cost-performance trade-off compared to baselines.
    \end{itemize}
\end{enumerate}

\section{Related Work}
\label{sec:related_work}

% \textcolor{red}{THIS SECTION IS BRAND NEW}
Our research is situated at the intersection of VLM-based autonomous agents, offline reinforcement learning, and inference-time value estimation. We classify prior work into three categories: VLM agents for web tasks, training-heavy policy improvement methods, and inference-time decision-making strategies.
\paragraph{VLM-based Agents for GUI Control.}
Vision-Language Models (VLMs) have emerged as powerful end-to-end policies for navigating digital environments. Early frameworks like WebVoyager \cite{he_webvoyager_2024} and Auto-GUI \cite{zhang_you_2024} demonstrated that VLMs can directly map screenshots and instructions to actions. CogAgent \cite{hong_cogagent_2024} further enhanced this by introducing high-resolution visual encoders for fine-grained GUI understanding. To improve the precise grounding of actions (e.g., specific coordinate clicks), models like SeeClick \cite{seeclick2024} have harnessed GUI grounding pre-training. More recent approaches, such as Agent-E \cite{abuelsaad_agent-e_2024} and UI-TARS \cite{qin_ui-tars_2025, wang_ui-tars-2_2025}, integrate hierarchical control and multi-turn RL to handle long-horizon tasks. However, achieving state-of-the-art performance with these architectures typically necessitates resource-intensive supervised fine-tuning on large-scale demonstration datasets, such as \textit{Android in the Wild} \cite{rawles_android_2023}. While effective, the performance of these agents remains bounded by the coverage of the training data and the immense computational cost of iteratively updating multi-billion parameter models.

\paragraph{Training-Based Policy Improvements.}
To improve agent performance beyond imitation learning, recent works have turned to Reinforcement Learning (RL). Online RL methods, such as MobileGUI-RL \cite{mobileguirl2025} and \cite{carta_grounding_2024} allow for policy updates via environment interaction but are often prohibitively slow for web tasks due to the need for live execution. Offline RL offers a paradigm to learn from static datasets, though it faces the challenge of \textit{distributional shift}. CCommon solutions involve policy constraints, such as ARPO \cite{ARPO2025} which utilizes experience replay for end-to-end optimization, or Advantage-Weighted Regression (AWR) \cite{AWR2019} used in DigiRL \cite{bai_digirl_2024}.
Recent research has prioritized stabilizing value estimation in reasoning tasks, as seen in methods like VAPO \cite{vapo2025}.
Concurrently, frameworks such as InSTA \cite{trabucco_insta_2025} and PAE \cite{zhou_proposer-agent-evaluatorpae_2024} have emerged to fuel training pipelines through autonomous data generation; these systems leverage LLMs to synthesize tasks and evaluate trajectories, thereby establishing self-improvement loops . Taking a different approach, Memento \cite{zhou_memento_2025} bypasses the need for direct parameter updates by utilizing a memory bank to retrieve past trajectories. A common drawback across these methods is their reliance on either updating the massive policy parameters or maintaining extensive external memory, both of which are computationally demanding.

\paragraph{Inference-Time Value Estimation and Ranking.}
Instead of retraining the policy, a growing body of work focuses on enhancing decision-making at inference time. Some approaches utilize the LLM itself as a value estimator or utilize verbal reinforcement to self-correct, as seen in Reflexion \cite{reflexion2023} and other critic-based methods \cite{ma_vision_2024, cao_beyond_2024}. Others apply heavy search algorithms to guide reasoning, such as Search-o1 \cite{searcho12025} or Monte Carlo Tree Search (MCTS) combined with step-level Q-value models \cite{zhai_enhancing_2024}, or heuristic search ($Q^*$) \cite{wang_q_2024}, to guide reasoning.
Our work is most closely related to approaches that learn offline value functions to guide frozen models. We leverage Implicit Q-Learning (IQL) \cite{kostrikov_offline_2021} to train a stable Q-function without querying out-of-distribution actions. This distinguishes our method from DigiQ \cite{bai_digi-q_2025}, which also learns a Q-function over VLM features but uses it to relabel data for subsequent \textit{policy retraining}. In contrast, our approach applies the Q-function directly for \textit{inference-time reranking}. By decoupling the high-capacity VLM proposer from a lightweight, specialized value network, we achieve immediate policy improvement without the computational overhead of fine-tuning or the latency of heavy search procedures.

\section{The Best-of-Q Agent}
\label{sec:best-of-q-agent}

\begin{figure*}[t] 
    \centering
    \includegraphics[width=\linewidth]{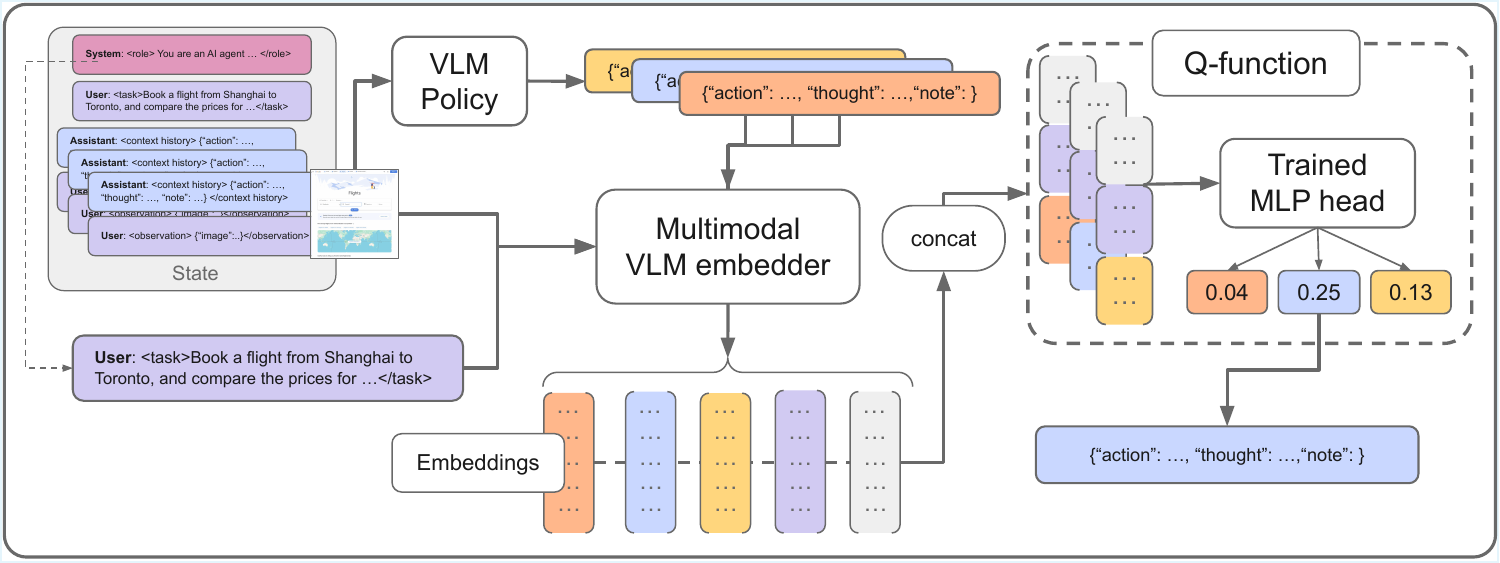}
    \caption{Structure of the Best-of-Q Agent. The multi-modal state goes through a VLM policy, prompted to give $N$ action candidates. A frozen VLM processes raw state information and those action candidates into separate embeddings. These fixed-size vectors are then concatenated and fed into a lightweight Multi-Layer Perceptron (MLP) which outputs the Q-value used for reranking candidate actions at inference time.}
    \label{fig:main-structure-boq}
\end{figure*}

Our approach, which we term the \textbf{Best-of-Q} agent, is centered on decoupling the agent's logic into two distinct components: (1) a frozen, high-capacity VLM that acts as an \textbf{action proposer}, generating a set of $N$ candidate actions for a given state, and (2) a lightweight, offline-trained \textbf{Q-function} that acts as an \textbf{action selector}. This selector reranks the $N$ candidates to execute the one with the highest predicted value. The overall structure of this agent is illustrated in Figure \ref{fig:main-structure-boq}.

\subsection{Problem Formulation} % Changed title for clarity
\label{sec:rl_framework_condensed}

Following prior work (\cite{he_webvoyager_2024}, \cite{yang_react_2024}), we model the web agent task as a Markov Decision Process (MDP). The state $s$ is multi-modal, including the task instruction, action history, and current screenshot. Actions $a$ consist of an action type (e.g., Click, Write) and associated reasoning. The environment dynamics are unknown, and the reward $R(s, a)$ is sparse: $1$ upon successful task completion, $0$ otherwise. Our objective is to learn an optimal policy $\pi$ that maximizes the expected discounted cumulative reward.

Central to our approach are the standard \textbf{state-value function} $V^{\pi}(s)$ and \textbf{action-value function} $Q^{\pi}(s,a)$:
\begin{equation}
    V^{\pi}(s) = \mathbb{E}_{\pi} \left[ \sum_{t=0}^{T} \gamma^t r_t | s_0 = s \right]
\end{equation}

\begin{equation}
    Q^{\pi}(s,a) = \mathbb{E}_{\pi} \left[ \sum_{t=0}^{T} \gamma^t r_t | s_0 = s, a_0 = a \right]
\end{equation}

We specifically aim to learn the action-value function $Q(s,a)$ via offline RL to serve as a scoring mechanism for ranking candidate actions at inference time.

\subsection{Offline Q-Function Training and Architecture}
\label{sec:offline-training-and-arch}

A robust Q-function must learn to differentiate between high- and low-quality actions, which requires training on a dataset with diverse outcomes. To ensure broad exploration of action candidates, we begin by gathering an initial dataset with an $\epsilon$-greedy policy. Figure \ref{fig:eps-greedy-agent} in Appendix describes that policy, showing how we can leverage the multi-action proposal design to explore with a given probability. We then train an initial Q-function on this data and use the resulting Best-of-Q Agent to collect new, higher-quality exploitative trajectories. This cycle is repeated, creating a virtuous cycle that progressively improves both the dataset $\mathcal{D}$ and the Q-function. See Appendix \ref{sec:iterative-data-collection} for additional details on this data collection strategy.

% \begin{figure*}[t]
%     \centering
%     \includegraphics[width=0.9\textwidth]{images/q-v functions.jpeg}
%     \caption{Architectural overview of the Best-of-Q Agent. A frozen VLM processes raw state information into separate embeddings. These fixed-size vectors are then concatenated and fed into a lightweight MLP which outputs the Q-value used for reranking candidate actions at inference time.}
%     \label{fig:architecture_diagram}
% \end{figure*}

Our architecture decouples perception from value estimation (see Fig. \ref{fig:main-structure-boq}). We use a frozen VLM as a feature extractor. For any given transition, it produces fixed-size embeddings for the state ($s_t$), action ($a_t$), and task separately. These embeddings are then fed into a Multi-Layer Perceptron (MLP with parameters $\theta(s, a)$) which serves as our Q-function, $Q_\theta(s, a)$, and is the only component we train. The detailed feature extractor and MLP architectures are available in Appendix \ref{sec:architecture}.

We train this lightweight Q-function using Implicit Q-Learning (IQL) \cite{kostrikov_offline_2021}. We chose IQL for its high stability in offline, sparse-reward environments, as it learns a value function without querying out-of-distribution (OOD) actions, thus avoiding distributional shift. First, we learn a state-value function $V(s)$ via expectile regression:

\begin{equation}
    L_{V}(\psi) = \mathbb{E}_{(s,a)\sim\mathcal{D}} \left[ L_{2}^{\tau}(Q_{\overline{\theta}}(s,a) - V_{\psi}(s)) \right],
\end{equation}

with $L_2^\tau(u) = |\tau - \mathbb{I}(u < 0)| u^2$, and using this $V(s)$ as a stable target to update the Q-function $Q(s,a)$:

\begin{equation}
\begin{split}
    L_{Q}(\theta) = \mathbb{E}_{(s,a,s^{\prime})\sim\mathcal{D}} \left[ \left(r(s,a) + \gamma V_{\psi}(s^{\prime}) \right. \right. \\
    \left. \left. - Q_{\theta}(s,a)\right)^2 \right].
\end{split}
\end{equation}
Here, $V_{\psi}(s)$ and $Q_{\theta}(s,a)$ are the state-value and Q-function networks, parameterized by $\psi$ and $\theta$ respectively, while $Q_{\overline{\theta}}$ is a slow-moving target network to stabilize training, parametrized by $\overline{\theta}$. Transitions $(s,a,r,s')$ are sampled from the static offline dataset $\mathcal{D}$. The expectile loss $L_{2}^{\tau}(u) = |\tau - \mathbb{I}(u < 0)| u^2$ uses a hyperparameter $\tau \in (0,1)$ to push $V_{\psi}(s)$ towards the maximum action-value seen for state $s$ in the dataset. Using the reward $r$, discount factor $\gamma$, and this learned state-value $V_{\psi}(s')$ as a stable target is the key mechanism in IQL, as it allows the Q-function to be updated without querying potentially out-of-distribution actions.
Additional details on those learning objectives are given in Appendix \ref{sec:iql-training}.

\subsection{Action Selection at Inference Time}

At inference time, the agent leverages the trained Q-function to guide its decision-making process. For every step, the policy generates a discrete set of $N$ action candidates. A value of 3 was used for $N$, and an ablation on the influence of this parameter is conducted in section \ref{sec:influence-of-n}. The Q-value model then scores each candidate, providing an estimated value of the future cumulative reward for that specific action. This process is analogous to classic RL algorithms like Deep Q-Networks (DQN \cite{mnih_playing_2013}), which select the action with the highest Q-value from a discrete action space. However, while DQN operates on a predefined, finite set of actions, our approach extends this principle to a generated subspace of the agent's infinite action space. We select the action with the maximum Q-value from the set of candidates, with the assumption that at least one of these candidates represents an optimal or near-optimal choice for the current state. By performing this \texttt{argmax} over the generated candidates, our agent can significantly improve its performance in complex multi-step tasks (Section \ref{sec:results}).

\section{Experiments}
\label{sec:experiments}

\begin{table}[t!]
\centering
\small % Shrinks font size to help it fit
\begin{threeparttable}
\caption{Dataset Statistics per VLM Policy} % Changed caption
\label{tab:dataset-stats} % Changed label
% Changed tabularx definition to 3 columns: |X|c|c|
\begin{tabularx}{\columnwidth}{|>{\raggedright\arraybackslash}X|c|c|}
\hline
\textbf{VLM Policy} & \textbf{Total Episodes} & \textbf{Success Rate (\%)} \\ % Changed headers
\hline
GPT4.1 & 9,060 & 82.0 \\ % Added comma for readability and fixed percentage format
\hline
Qwen2.5-72B & 17,250 & 71.8 \\ % Added comma
\hline
Qwen2.5-7B & 15,112 & 48.4 \\ % Added comma
\hline
\end{tabularx}
\end{threeparttable}
\end{table}

\begin{table*}[t!]
\centering
\begin{threeparttable} % Wraps everything to align notes with table width
    \caption{WebVoyager Performance and Efficiency Comparison}
    \label{tab:webvoyager_results}
    
    % The tabularx takes the full textwidth
    \begin{tabularx}{\textwidth}{|>{\raggedright\arraybackslash}X|>{\raggedright\arraybackslash}X|c|c|}
        \hline
        \textbf{VLM Agent Backbone} & \textbf{Method} & \textbf{Success Rate ($\uparrow$)} & \textbf{Avg. Steps for Success ($\downarrow$)} \\
        \hline
        \multicolumn{4}{|c|}{\textit{Proprietary Models}} \\
        \hline
        \texttt{GPT-4.1} & Prompting & 82.4 $\pm$ 1.0\% & 7.0 $\pm$ 0.2 \\
        \texttt{GPT-4.1} & + Random Action & 72.1 $\pm$ 1.5\% & 11.3 $\pm$ 0.3 \\
        \texttt{GPT-4.1} & + \textbf{Best-of-Q} & \textbf{88.8$\pm$ 0.9\%} & \textbf{7.1 $\pm$ 0.2} \\
        \hline
        \multicolumn{4}{|c|}{\textit{Open-Source Models}} \\
        \hline
        \texttt{Qwen2.5-VL-7B} & Prompting & 38.8 $\pm$ 1.2\% & 10.9 $\pm$ 0.4 \\
        \texttt{Qwen2.5-VL-7B} & + Random Action & 40.3 $\pm$ 1.4\% & 11.9 $\pm$ 0.4 \\
        \texttt{Qwen2.5-VL-7B} & + \textbf{Best-of-Q} & \textbf{55.7 $\pm$ 1.3 \%} & \textbf{10.3 $ \pm$ 0.3} \\
        \hline
        \texttt{Qwen2.5-VL-72B} & Prompting & 69.1 $\pm$ 1.3\% & 9.4 $\pm$ 0.2 \\
        \texttt{Qwen2.5-VL-72B} & + Random Action & 58.3 $\pm$ 1.4\% & 12.0 $\pm$ 0.3 \\
        \texttt{Qwen2.5-VL-72B} & + \textbf{Best-of-Q} & \textbf{76.6 $\pm$ 1.0\%} & \textbf{8.8 $ \pm$ 0.2} \\
        \hline
    \end{tabularx}
    
    \begin{tablenotes}
        \small
        \centering
        \item \textbf{Success Rate ($\uparrow$)}: The percentage of tasks successfully completed. Higher is better.
        \item \textbf{Avg. Steps for Success ($\downarrow$)}: The average number of steps taken to complete successful tasks. Lower is better.
        \item \textbf{Method}: Our method (\textbf{+ Best-of-Q (Ours)}) is compared against standard prompting and other offline RL baselines.
    \end{tablenotes}
\end{threeparttable}
\end{table*}

\subsection{Setup}

We evaluated our method on the WebVoyager benchmark \cite{he_webvoyager_2024, hong_cogagent_2024}. This benchmark is composed of 643 tasks, on 15 different domains. Since the tasks are performed by the agent on real websites, on the live internet, the original benchmark is not valid anymore (due to dates in the past, websites changing, products that do not exist anymore, etc). Following the guidelines associated with the benchmark's leaderboard, we used the set of 590 tasks patched by the SOTA holders \footnote{\url{https://github.com/sagekit/webvoyager/blob/main/data/patchedTasks.jsonl}}. The count and domains are provided in Appendix \ref{sec:webvoyager-details}. We compare our method, called \textbf{Best-of-Q}, to two baseline agents:
\begin{itemize}
    \item \textbf{Prompting} The VLM policies are prompted to solve the task, and to give only one action at each step.
    \item \textbf{Random action} This is the $\epsilon$-greedy agent with $\epsilon = 1$, i.e. an agent that proposes $N$=3 actions at each step, and one is taken at random to be executed. 
\end{itemize}
We compare different VLMs as policies, from proprietary (GPT4.1) to open-sourced (Qwen2.5-VL) of different sizes.
To asses the superiority of one agent, we compare both the success rate on the benchmark (averaged on 3 runs) and the average number of steps it takes to successfully achieve a task.

\subsection{Training samples and hyperparameters}
\label{sec:training-samples-and-hyperparams}

As described in Section \ref{sec:offline-training-and-arch} and in Appendix \ref{sec:iterative-data-collection}, we collected data iteratively in order to train a Q-function with IQL. For all VLM policies compared, Table \ref{tab:dataset-stats} references the amount of episodes and corresponding success rate of the offline datasets used. 

The main hyperparameter for training is the expectile $\tau$ parameter of the expectile loss of IQL. To train on data collected with the GPT4.1 policy, the expectile-$\tau$ was set at 0.8, and for the other 2 datasets it was lowered at 0.7 since the datasets were composed of many more failures and a high $\tau$ was leading to an over confident Q value, and poorer results. More details about hyperparameters are available in Appendix \ref{sec:hyperparameters-training}.

\subsection{Results}
\label{sec:results}

Our results, summarized in Table~\ref{tab:webvoyager_results}, demonstrate that the \textbf{Best-of-Q} method is model-agnostic and significantly enhances agent performance across all VLM backbones. The approach is effective even on powerful proprietary models, boosting GPT-4.1 from 82.4\% to \textbf{88.8\%} (+6.4\%), a clear gain for a model that cannot be fine-tuned. The method is particularly impactful on open-source models, turning the modest \texttt{Qwen2.5-VL-7B} into a much more capable agent by lifting its success rate from 38.8\% to \textbf{55.7\%} (+16.9\%). It also improves the \texttt{Qwen2.5-VL-72B} from 69.1\% to \textbf{76.6\%} (+7.5\%), while simultaneously increasing its efficiency (reducing average steps from 9.4 to 8.8).

Crucially, the poor performance of the \textbf{Random Action} baseline (which dropped \texttt{GPT-4.1} to 72.1\%) proves that the success of our method is not merely due to having more action candidates. Instead, it underscores the importance of a principled and informed action-selection mechanism. This suggests that a primary cause of VLM suboptimality is its flawed greedy selection, which our lightweight Q-function effectively corrects. A detailed per-domain breakdown of these results is available in Appendix \ref{sec:extended-results}. We propose an analysis on \textit{why} our method outperforms the prompting baseline in Appendix \ref{sec:variance_analysis} diving into variance analysis on tasks completion.
\subsection{Cost analysis}
\label{sec:cost-analysis}

Another way to benchmark Best-of-Q against baselines is to evaluate the total cost of a full run. Specifically, we aggregate the cost of the total input and output tokens generated by the VLM policy, and we add (if present) the estimated cost of the Q-function calls. We limit those to the costs of calling the VLM embedder in terms of input tokens, neglecting the output cost since we do not generate any tokens. Indeed, we only output a single numeric value through the Q-function MLP, and its size is significantly small in comparison to the rest. We plot the cost of the agent against its performance on the benchmark in Figure \ref{fig:pareto-cost-perf}. Details on values used to compute those results are provide in Appendix \ref{sec:cost-details}.

\begin{figure}[t!]
    \centering
    \includegraphics[width=0.95\linewidth]{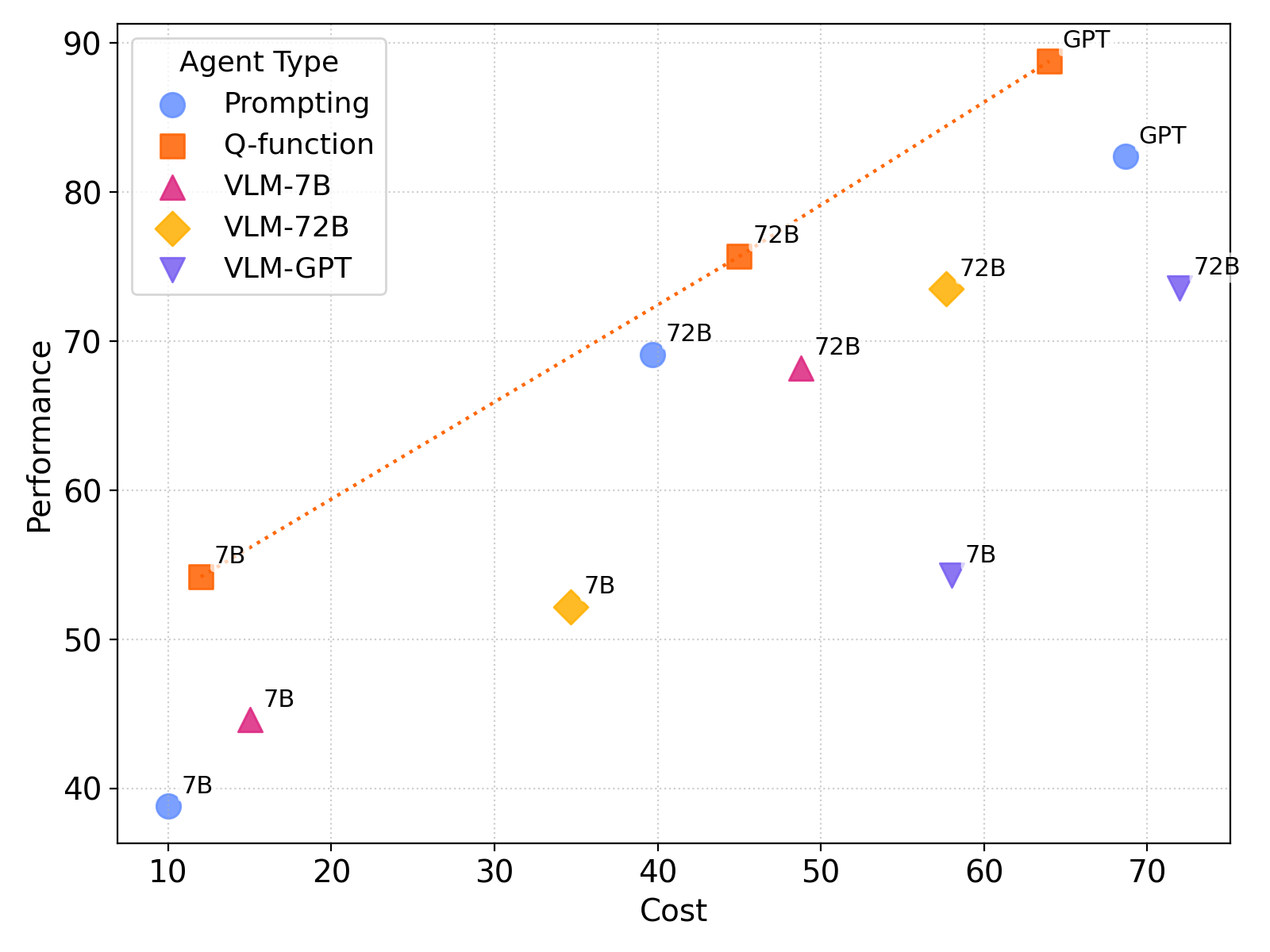}
    \caption{Pareto plot of the Best-of-Q Agent against prompting baselines, and VLM-action selector baselines. The cost is in USD/benchmark averaged on 3 runs. The points associated to "VLM-7B", "VLM-72B" and "VLM-GPT" are references to agents using a VLM as an action selector in place of a trained Q-function. For further insights on those agents and that study, refer to Section \ref{sec:vlm-as-sampler}.}
    \label{fig:pareto-cost-perf}
\end{figure}

\subsection{Failure mode analysis}
\label{sec:failure_modes}

While our method significantly boosts the performance of the base VLM agent, it does not achieve a 100\% success rate. This remaining performance gap can be attributed to two primary failure modes: 1) a limited action-proposal ability in the VLM policy, and 2) an imperfect Q-function that struggles to choose the optimal action from the available candidates.

We hypothesize that the agent's performance is fundamentally capped by the quality of the actions proposed by the base VLM. Our Q-function, while effective at reranking, cannot select an action that was never proposed. To gather evidence supporting this hypothesis, we conducted a granular, step-level analysis on the challenging Google Flights subdomain of WebVoyager.

\begin{figure}[t]
    \centering
    \includegraphics[width=0.9\linewidth]{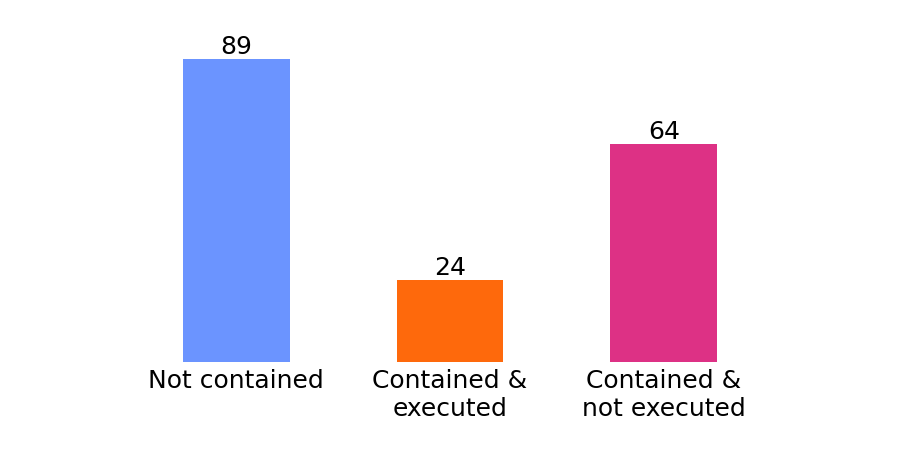}
    \caption{Breakdown of agent performance on a challenging Google Flights task. The analysis shows that in 50.2\% of steps, the base VLM (Qwen2.5-VL-7B) fails by not proposing the correct "golden" action (as defined by a stronger GPT-4.1 policy). When the correct action is proposed, our Q-function successfully selects it in 13.6\% of cases but fails to do so in the remaining 36.2\%. This demonstrates that the primary performance bottleneck is the VLM's limited proposal ability, not the Q-function's selection process}
    
    \label{fig:failure_breakdown}
\end{figure}

To diagnose these failure modes, we conducted a step-level analysis on a complex task from the Google Flights subdomain: \textit{Find flights from Chicago to London on 20 December and return on 23 December}. Our agent (using Qwen2.5-VL-7B) consistently fails this task, whereas an agent using GPT-4.1 consistently succeeds. We used the successful GPT-4.1 trajectory as a "gold standard" to identify the correct action at each step. We then analyzed whether our agent's base VLM ever proposed this correct action within its candidate set. The prompt used for this analysis is in Appendix \ref{sec:vlm-selector-prompt}.
The data presented in Figure \ref{fig:failure_breakdown} clearly shows that the system's primary bottleneck is the VLM's action-proposal ability. The single largest slice of the pie chart, at 50.3\%, represents transitions where the correct "golden action" was not contained in the Qwen2.5-VL-7B's list of candidates. In these cases, no matter how effective our Q-function is, it cannot select a correct (or at least optimal) action because none were offered.

When a correct action was present in the candidate list, our Q-function successfully selected it in 13.6\% of cases. However, in the remaining 36.2\% of these instances, the Q-function failed to select the correct action, highlighting its own imperfection. These failures may stem from factors such as value ambiguity between similar actions or insufficient coverage for specific state-action pairs in the offline training data. While we acknowledge that using a single "golden" trajectory may not account for all viable paths, the evidence overwhelmingly points to a primary bottleneck. The instances of Q-function error (36.2\%) are significantly outnumbered by the cases where the VLM fails to propose the correct action in the first place. This provides strong quantitative support for our central hypothesis: the agent's performance is fundamentally constrained by the VLM's proposal mechanism, and while our Q-function provides significant gains, its efficacy is ultimately capped by the quality of the candidate actions it receives.

A complementary experiment using a GPT-4.1 sampler can be found in Section \ref{sec:vlm-as-sampler} as well as a qualitative study of trajectories in Appendix \ref{sec:qualitative-study}.

\subsection{Discussion on Specialized Fine-Tuning}
\label{sec:holo-comparison}

A direct comparison with fine-tuning baselines was not feasible due to computational constraints. Representative offline RL methods such as Filtered Behavioral Cloning (FBC), which imitates only high-performing trajectories, and Advantage-Weighted Regression (AWR, \cite{AWR2019}), which re-weights training samples based on their estimated advantage, both necessitate a full fine-tuning pass on the VLM policy.

However, to provide a performance proxy for a highly optimized, resource-intensive fine-tuned agent, we compare our results to Holo1-7B \cite{andreux_surfer-h_2025}, a specialized model based on the same Qwen2.5-7B backbone. We must emphasize that \textbf{this is not a direct baseline}. The Holo1-7B model was trained via a sophisticated multi-stage pipeline, including Filtered Behavioral Cloning (FBC) and performance distillation from a much larger policy (e.g., Qwen 32B), using a large-scale dataset of high-quality (around 1.5 million), successful trajectories. In contrast, our Q-function was trained on a much smaller (around 300000 transitions), mixed-quality dataset collected via an $\epsilon$-greedy policy. An FBC baseline trained on our data would be at a significant disadvantage and would not achieve comparable results, as our dataset contains far fewer successful traces than the high-quality, expert-level data used to train Holo1.

The specialized Holo1-7B agent achieves a success rate of $59.2\% \pm 1.2\%$ on our benchmark setup. Our Best-of-Q Agent, using the same 7B backbone but without any VLM fine-tuning, achieves a highly competitive $55.7\% \pm 1.3\%$. This result is significant: our method, which only requires training a lightweight ($\sim$11M parameter) Q-function on suboptimal data, delivers performance in a similar range to a fully fine-tuned, specialized agent trained on expert data. This demonstrates a superior trade-off between computational cost and performance.

\subsection{Sample efficiency}
\label{sec:sample-efficiency}
\begin{figure*}[t!]
    \centering
    \begin{subfigure}[b]{0.49\linewidth}
        \includegraphics[width=\linewidth]{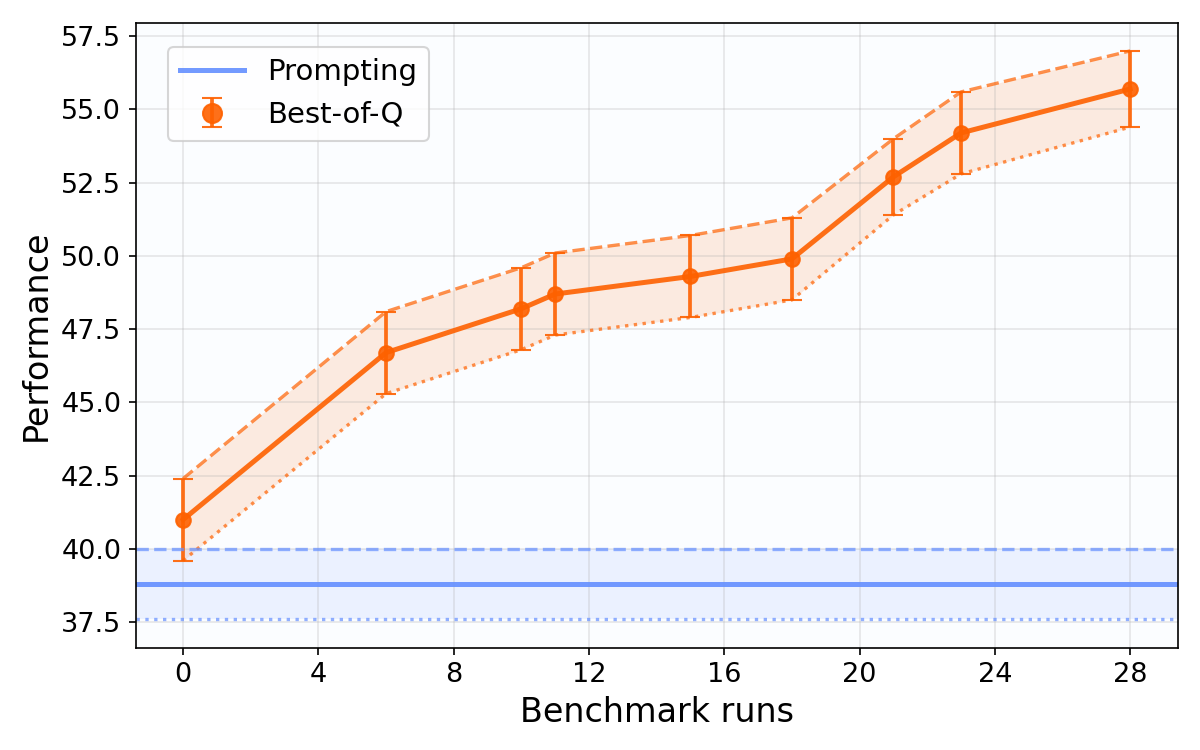}
        \caption{Qwen2.5-VL-7B policy.}
        \label{fig:sample-efficiency-7b}
    \end{subfigure}
    \hfill
    \begin{subfigure}[b]{0.49\linewidth}
        \includegraphics[width=\linewidth]{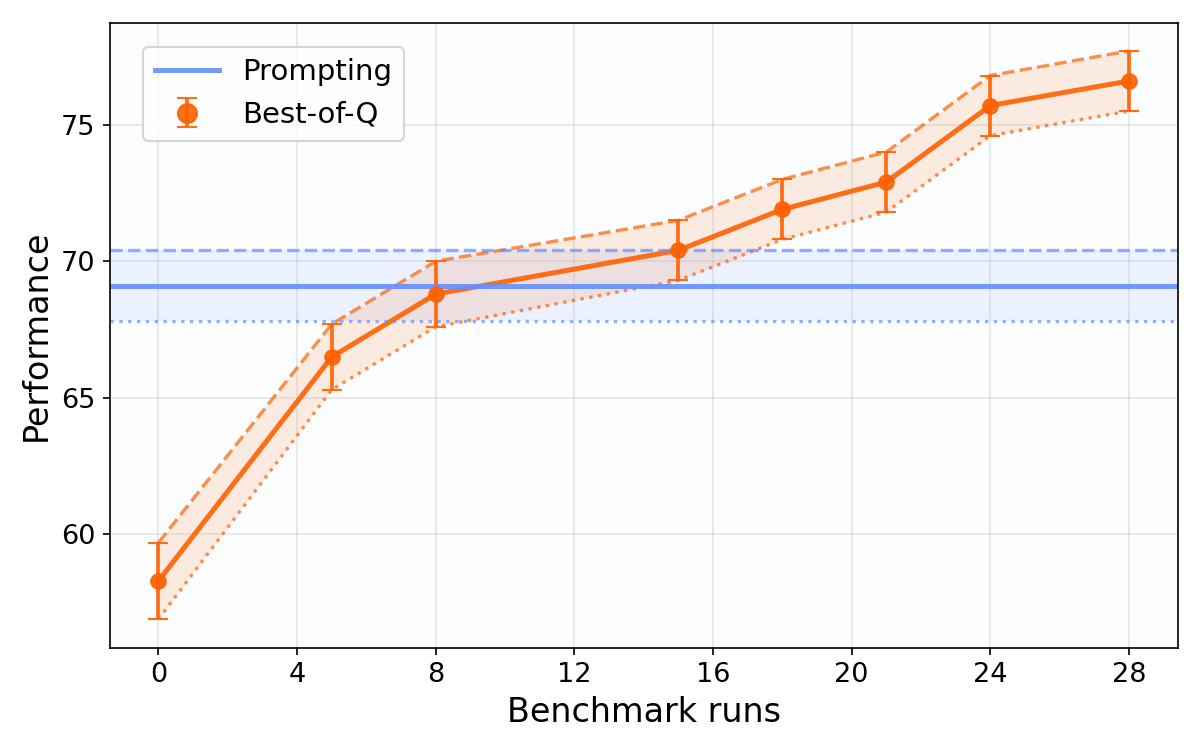}
        \caption{Qwen2.5-VL-72B policy.}
        \label{fig:sample-efficiency-72b}
    \end{subfigure}
    \caption{Performance of the Best-of-Q Agent as a function of the number of trajectories used for Q-function training. We see that the agents are self-improving the more data it generated and on which it is trained.}
    \label{fig:sample-efficiency}
\end{figure*}
Sample efficiency in reinforcement learning refers to how well an algorithm learns a task relative to the number of interactions, or "samples," it requires with the environment. A more sample-efficient algorithm achieves strong performance with less data, which is crucial for real-world applications where data collection is often costly and time-consuming. We analyze the sample efficiency of our Best-of-Q method by evaluating the evolution of our agent's performance as we incrementally increase the size of the training dataset. Figure \ref{fig:sample-efficiency} illustrates the performance of our Best-of-Q Agent for the Qwen2.5-VL-7B and Qwen2.5-VL-72B policies, with the horizontal blue line indicating the baseline performance of the standard prompting agent. The x-axis represents the cumulative number of runs (trajectories) added to the training dataset.

As shown in Figure \ref{fig:sample-efficiency-7b}, the Best-of-Q Agent using the Qwen2.5-VL-7B policy demonstrates a clear improvement over the baseline after just a single training run. The performance curve shows a steep initial increase, with the agent's success rate surpassing the baseline within the first few runs. This suggests that the Q-function is highly sample-efficient, quickly learning to identify and exploit valuable actions from a small amount of data. Subsequent training iterations lead to continued performance gains, eventually reaching a plateau around 27 runs.

In Figure \ref{fig:sample-efficiency-72b}, we see a similar pattern for the more capable Qwen2.5-VL-72B policy. The agent's performance exceeds the baseline almost immediately. The curve demonstrates a consistent upward trend, but unlike the 7B model, its performance gain is less dramatic relative to the high-performing baseline. This is expected, as the 72B model's inherent reasoning capabilities are already very strong, leaving less room for the Q-function to improve upon the base policy.

The analysis confirms that our iterative data collection and training approach is highly effective. It creates a virtuous cycle: the agent, guided by an increasingly accurate Q-function, collects progressively better data, which in turn is used to train an even more capable Q-function. The results show that our method is remarkably sample-efficient, requiring only a modest number of trajectories to achieve significant and consistent performance improvements over the baseline. 

\section{Ablations}
\label{sec:ablations}
We propose ablations around key elements of our approach. Those are the importance of 1) the Q-function itself, 2) the influence of the number of action candidates, and 3) the influence of the VLM embedder of the Q-function. Details about the last ablation can be found in Appendix \ref{sec:influence-embedder}.
\paragraph{\textbf{Using a VLM as a Q-function}}
\label{sec:vlm-as-sampler}

\begin{table}[t] 
\centering
\small % Shrinks font size to help it fit
\begin{threeparttable}
\caption{Using a VLM as a Q-function action picker}
\label{tab:vlm-picker-results}
% 1. Changed \textwidth to \columnwidth
% 2. Changed 5-column definition to 4-column: |X|l|c|c|
\begin{tabularx}{\columnwidth}{|>{\raggedright\arraybackslash}X|l|c|c|}
\hline
\multicolumn{2}{|c|}{} & \multicolumn{2}{c|}{\textbf{Policy Model}} \\
\hline
% Shortened headers to fit single column
\multicolumn{2}{|l|}{\textbf{VLM Picker \textbf{$\downarrow$} } } & \textbf{Qwen2.5-72B} & \textbf{Qwen2.5-7B} \\
\hline
\textbf{GPT4.1} & Success ($\uparrow$) & 73.6 $\pm$ 1.1 \% & 54.3 $\pm$ 1.4\% \\
& Avg. Steps ($\downarrow$) & 8.7 $\pm$ 0.2 & 10.1 $\pm$ 0.3 \\
\hline
\textbf{Qwen2.5-72B} & Success ($\uparrow$) & 73.5 $\pm$ 1.1\% & 52.2 $\pm$ 1.4\%  \\
& Avg. Steps ($\downarrow$) & 9.2 $\pm$ 0.2 & 10.4 $\pm$ 0.3 \\
\hline
\textbf{Qwen2.5-7B} & Success ($\uparrow$) & 68.2 $\pm$ 1.1\% & 44.6 $\pm$ 1.1\% \\
& Avg. Steps ($\downarrow$) & 9.4 $\pm$ 0.2 & 9.9 $\pm$ 0.3 \\
\hline
\end{tabularx}
\end{threeparttable}
\end{table}

A simple yet more expansive way of picking an action at each step, would be to present the candidates, along with the state, to a VLM and ask it to provide the best option, alongside a reasoning to enforce and stabilize the answer. This experiment can be conducted with any VLM as an action selector replacing the Q-function completely. This has a major advantage relying in the fact that we don't need additional training whatsoever, but it relies on calling a large VLM a second time. The prompt given to a VLM to select the best action is provided in Appendix \ref{sec:vlm-selector-prompt}.  Results are reported in Table \ref{tab:vlm-picker-results}. Comparing with the main results in Section \ref{sec:results}, we can see that the Best-of-Q Agent outperforms all of those agents, with a smaller cost (see Section \ref{sec:cost-analysis}). That being said, those agents still improved on the simple prompting baseline, showing that providing alternatives is an effective way of boosting the planning and action selection capabilities of a VLM in the framework of web agents. 

\paragraph{\textbf{Influence of then number of actions $N$}}
\label{sec:influence-of-n}
We analyze the effect of changing the number of candidate actions ($N$) at inference time, using Q-functions trained on datasets collected with a fixed $N$. Table \ref{tab:influence-of-n-merged} compares two Q-functions: one trained on data collected with $N=3$ and another on data collected with $N=5$.

The results show that performance is maximized when the inference-time $N$ matches the training-time $N$ (bolded values in the table). For the $N=3$ trained model, the success rate peaks at \textbf{55.7\%} ($N=3$) and drops when $N$ is increased to 5 or 8. Similarly, the $N=5$ trained model achieves its best performance of \textbf{53.4\%} at $N=5$. Using an inference $N$ different from the one used for data collection does not improve, and often degrades, performance. This is likely due to two main factors:
\begin{enumerate}[leftmargin=*, nosep]
    \item \textbf{Distributional Shift:} Actions proposed with an $N$ larger than what the Q-function was trained on are effectively out-of-distribution, preventing accurate evaluation.
    \item \textbf{Low-Quality Candidates:} Large $N$ often results in "placeholder" actions (e.g., \textsc{Refresh}, \textsc{Restart}) that dilute the pool of viable options rather than adding strategic depth.
\end{enumerate}

\begin{table}[t] 
\centering
\small % Shrinks font size to help it fit
\begin{threeparttable}
\caption{Ablation on Inference-Time $N$}
\label{tab:influence-of-n-merged}
\begin{tabularx}{\columnwidth}{|l|l|>{\centering\arraybackslash}X|>{\centering\arraybackslash}X|}
\hline
 & & \multicolumn{2}{c|}{\textbf{Q-Function Trained on:}} \\
\cline{3-4}
% 2. Made first header bold to match the rest
\textbf{Inference N} & \textbf{Metric} & \textbf{N=3 Data} & \textbf{N=5 Data} \\
\hline
\textbf{N=3} & Success ($\uparrow$) & \textbf{55.7 $\pm$ 1.4\%} & 53.1 $\pm$ 1.4\% \\
& Steps ($\downarrow$) & \textbf{9.7 $\pm$ 0.3} & 9.6 $\pm$ 0.3 \\
\hline
\textbf{N=5} & Success ($\uparrow$) & 53.9 $\pm$ 1.1\% & \textbf{53.4 $\pm$ 1.4\%} \\
& Steps ($\downarrow$) & 10.3 $\pm$ 0.3 & \textbf{10.9 $\pm$ 0.4} \\
\hline
\textbf{N=8} & Success ($\uparrow$) & 54.5 $\pm$ 1.4\% & 52.1 $\pm$ 0.8\% \\
& Steps ($\downarrow$) & 13.4 $\pm$ 0.4 & 12.6 $\pm$ 0.4 \\
\hline
\end{tabularx}
\begin{tablenotes}
    \item Note: Bolded values are where inference $N$ matches the $N$ used during data collection.
\end{tablenotes}
\end{threeparttable}
\end{table}

\section{Conclusion}
\label{sec:conclusion}
In this work, we introduced the Best-of-Q Agent, a novel and efficient paradigm that enhances the performance of Vision-Language Model (VLM) agents without the high cost of policy fine-tuning. By decoupling the VLM's role as a powerful action proposer from the final action selection mechanism, we demonstrated a scalable solution that improves agent performance directly at inference time. Our method leverages a lightweight, offline-trained Q-function to rerank a set of candidate actions, enabling the agent to make more informed, value-based decisions at each step.

Through experiments, our approach proved to be a powerful and model-agnostic solution. We showed that the Q-function consistently boosted the success rates of both proprietary models like GPT-4.1 and open-source models like Qwen2.5-VL, turning suboptimal base policies into significantly more capable agents. The effectiveness of our method was particularly evident on the smaller Qwen2.5-VL-7B model, where our Q-function boosted its success rate by over 15 percentage points. Critically, this performance improvement was achieved with minimal computational overhead, as our cost analysis clearly showed a superior trade-off between performance and cost compared to standard prompting and random action baselines.

While our method significantly improves performance, our analysis indicates that the agent's success is primarily bounded by the recall of the base VLM's action proposals. Since the Q-function can only select from the generated candidates, instances where the VLM fails to propose a viable action represent a hard ceiling. Our quantitative breakdown confirms this is the dominant failure mode, highlighting a clear opportunity for future research, suggesting that combining our value-based action selection with methods to improve VLM proposal diversity or quality could lead to even greater improvements.

 \section*{Acknowledgements}

We would like to thank H Company for providing the infrastructure and resources that enabled this project, as well as Ludovic Denoyer for his vision and invaluable help. 

% \section*{Impact Statement}

% This work addresses the prohibitive cost of fine-tuning large Vision-Language Models (VLMs) for autonomous web agents by demonstrating that a lightweight, offline-trained Q-function can significantly enhance frozen model performance without parameter updates. By decoupling action ranking from policy generation, our framework democratizes access to state-of-the-art decision-making capabilities, bridging the gap between open-source and proprietary models for developers with limited compute resources. Furthermore, this approach offers substantial environmental benefits by eliminating the energy-intensive retraining cycles typical of RL or behavioral cloning methods, relying instead on a negligible MLP overhead. The method also enhances deployment safety by stabilizing stochastic VLM outputs through value-based reranking, reducing variance and hallucinations in critical workflows. While improving agent efficiency, we acknowledge the dual-use risks associated with scalable automation, such as automated account creation or CAPTCHA solving, underscoring the need for continued research into guardrails for autonomous systems interacting with live web environments.


\bibliography{Biblio-boq}
\bibliographystyle{icml2025}

%%%%%%%%%%%%%%%%%%%%%%%%%%%%%%%%%%%%%%%%%%%%%%%%%%%%%%%%%%%%%%%%%%%%%%%%%%%%%%%
%%%%%%%%%%%%%%%%%%%%%%%%%%%%%%%%%%%%%%%%%%%%%%%%%%%%%%%%%%%%%%%%%%%%%%%%%%%%%%%
% APPENDIX
%%%%%%%%%%%%%%%%%%%%%%%%%%%%%%%%%%%%%%%%%%%%%%%%%%%%%%%%%%%%%%%%%%%%%%%%%%%%%%%
%%%%%%%%%%%%%%%%%%%%%%%%%%%%%%%%%%%%%%%%%%%%%%%%%%%%%%%%%%%%%%%%%%%%%%%%%%%%%%%
\newpage
\appendix
\onecolumn
\section*{Appendix}

\section{More details on the Best-of-Q Agent}
\subsection{Actions with Localization}
\label{sec:localization}
Certain actions, such as \texttt{Click} or \texttt{WriteElement}, necessitate precise coordinates. Non-specialized models often struggle to simultaneously generate the correct high-level action and the exact pixel location required for execution. On the web, click operations frequently need to be highly precise (e.g., closing a small pop-up or selecting a specific date on a calendar). To address this, the agent makes an additional call to a dedicated \textbf{Localizer} (an internal specialized VLM) whenever precise coordinates are required. In our implementation, this model is an open-source Holo1.5-7B model, which has been fine-tuned on a diverse set of localization data, specifically for web interfaces.

\subsection{Data Collection via Iterative Policy Refinement}
\label{sec:iterative-data-collection}
Our approach to offline reinforcement learning begins with collecting a diverse dataset using an initial exploration policy. A robust Q-function requires a dataset containing both successful and unsuccessful examples to effectively learn to distinguish high-value actions from low-value ones. Relying on a purely greedy agent that proposes only a single action would produce a narrow, on-distribution dataset, severely limiting the Q-function's ability to generalize.

\begin{figure}[h]
    \centering
    \includegraphics[width=0.5\linewidth]{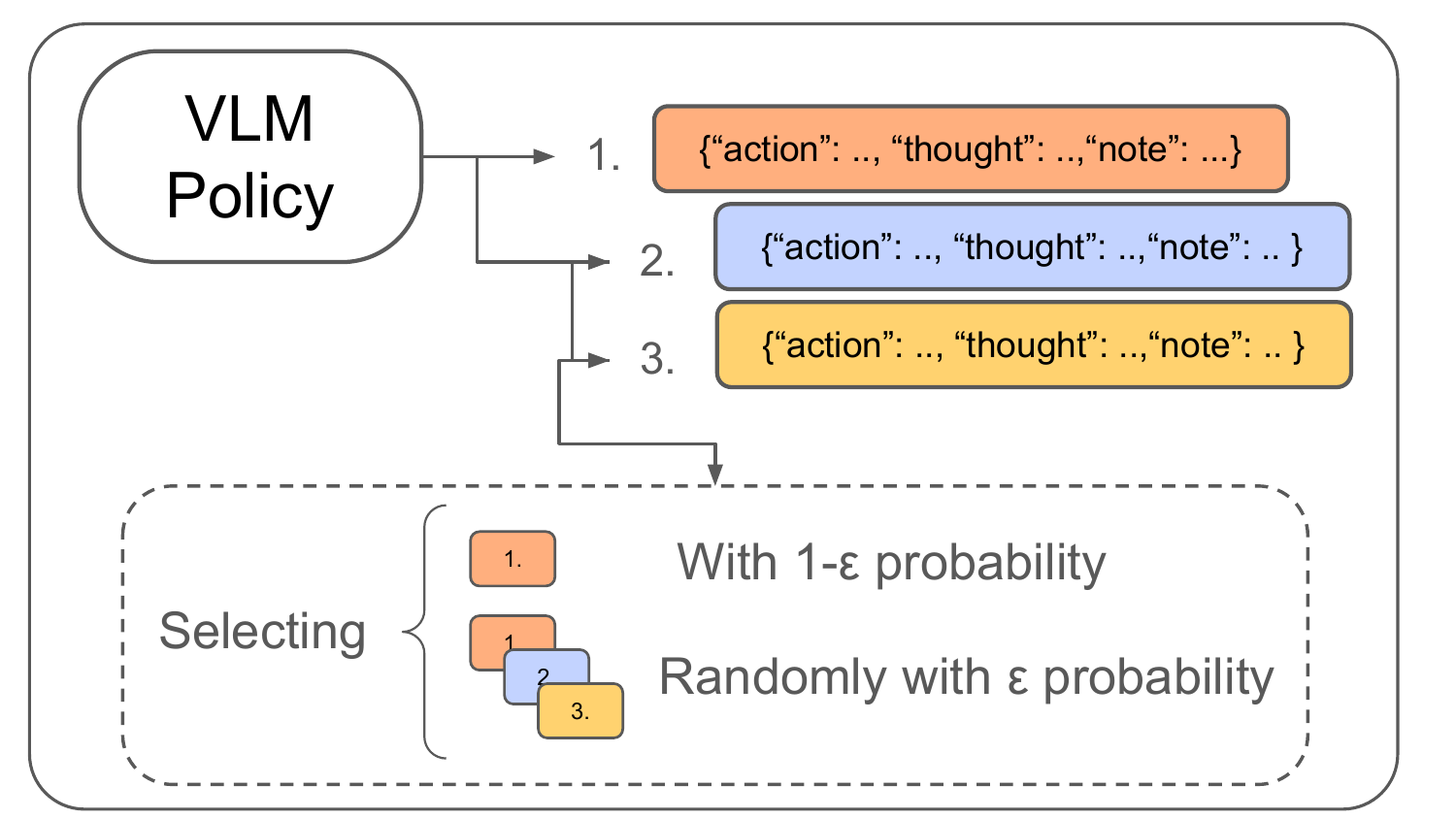}
    \caption{The $\epsilon$-greedy agent used for data collection. A typical value of $\epsilon$ ranges between 0.3 and 0.5.}
    \label{fig:eps-greedy-agent}
\end{figure}

\paragraph{Initial Exploration Phase.} To build a broad initial dataset, we leverage the VLM's ability to generate multiple candidate actions for any given state. While the first candidate typically represents the VLM's greedy choice, the others provide a rich source for exploration. We employ an $\epsilon$-greedy exploration policy, illustrated in Figure \ref{fig:eps-greedy-agent}, which selects the top-ranked action with probability $1-\epsilon$ and a random action from the remaining candidates with probability $\epsilon$. For our initial data collection, we set $\epsilon = 0.5$, balancing exploitation of the base policy's knowledge with exploration of its proposed alternatives. This phase generates a diverse "seed" dataset covering a wide range of state-action pairs.

\begin{figure*}[t]
    \centering
    \includegraphics[width=0.9\linewidth]{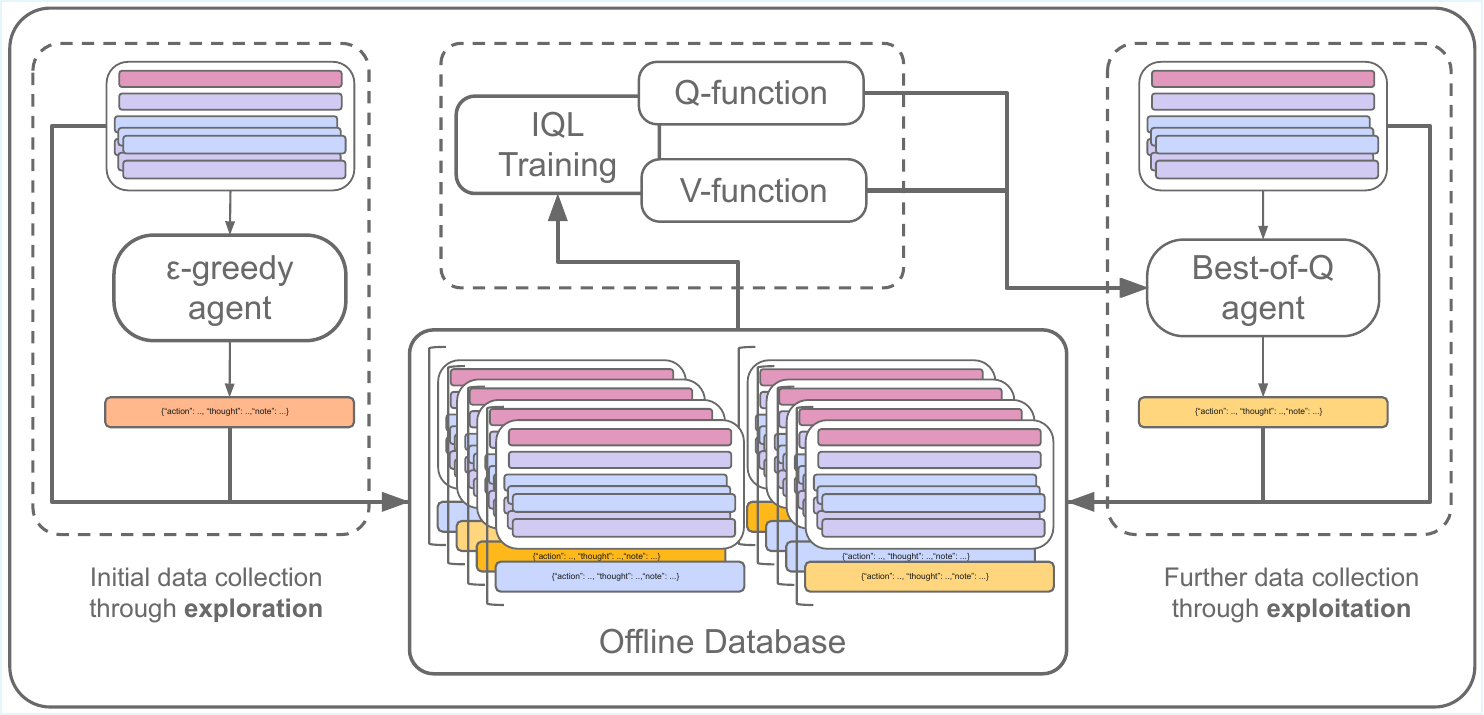}
    \caption{Data collection flow diagram. After the initial data collection with the $\epsilon$-greedy agent (Fig. \ref{fig:eps-greedy-agent}), a first Q-function is trained with IQL (Section \ref{sec:iql-training}). This Q-function is then used in the Best-of-Q Agent to collect additional data iteratively.}
    \label{fig:data-collection-flow}
\end{figure*}

\paragraph{Iterative Data Augmentation and Retraining.} A key element of our methodology is an iterative, multi-turn process of data collection and model retraining, inspired by self-improvement paradigms seen in works like WebRL \cite{qi_webrl_2025}. After the initial exploration phase, we train a preliminary Q-function on this "seed" dataset. This trained Q-function is then used to guide a new round of data collection, where the agent operates in a fully greedy, exploitative mode, always selecting the action with the highest predicted Q-value. These new, higher-quality trajectories are then added to our training data pool.

This cycle of \textbf{collect $\rightarrow$ train $\rightarrow$ exploit $\rightarrow$ add data $\rightarrow$ retrain} is repeated several times. In our experiments, we typically perform an initial collection of five runs, followed by four iterative cycles where two new exploitative runs are added to the training mixture in each cycle. This iterative refinement process creates a virtuous cycle: the agent, guided by an increasingly accurate Q-function, collects progressively better data, which in turn is used to train an even more capable Q-function. This process yielded our final dataset of 300,000 transitions with varying success distributions depending on the agent’s VLM policy, as described in Section \ref{sec:training-samples-and-hyperparams}. The final performance of our agent, as reported in Section \ref{sec:results}, is the result of this multi-stage training process. A supplementary analysis showing the performance evolution at each stage of this iterative process is included in Section \ref{sec:sample-efficiency}, demonstrating that the method avoids self-confirmation bias by continuously learning from new experiences as the dataset expands.

\subsection{Architectural Design}
\label{sec:architecture}
Our architecture is predicated on the principle of \textbf{decoupling perception from value estimation}. We leverage a large, pre-trained Vision-Language Model (VLM) for its powerful perceptual capabilities while offloading fine-grained decision-making to a lightweight, specialized Q-function. This design achieves significant policy improvement without the prohibitive cost of fine-tuning the multi-billion-parameter VLM.

\paragraph{VLM as a Frozen Feature Extractor.}
The foundation of our architecture is a pre-trained, frozen VLM (specifically Qwen2.5-VL-3B-Instruct). Its role is not to act as the policy, but as a high-fidelity feature extractor. For any given multi-modal state (composed of screenshots, task descriptions, and action history), the VLM processes the raw data through its transformer layers. To create a fixed-size representation suitable for a downstream network, we take the VLM's final hidden-layer outputs and apply a pooling operation (e.g., mean pooling). This step distills the rich, high-dimensional output into a compact embedding vector. This process is applied independently to generate three distinct embeddings for each transition: a \textbf{state embedding} ($s_t$), an \textbf{action embedding} ($a_t$), and a \textbf{task embedding}. Separating these representations allows our value model to learn the specific contribution of each component to the final predicted value.

\paragraph{Lightweight Value and Q-Function Networks.}
The core of our learning framework is a small Multi-Layer Perceptron (MLP) that serves as the Q-function, $Q_\theta(s, a)$. This network is designed for computational efficiency and is the only component with trainable parameters. The architecture first projects the state, action, and task embeddings into a shared dimension before concatenating them. This concatenated vector is then passed through a feed-forward network with several hidden layers, using ReLU activation functions and Dropout for regularization. The final layer outputs a single scalar value representing the predicted Q-value.

A typical configuration for our Q-network, which results in approximately 10 million parameters, is structured as follows:
\begin{enumerate}
    \item Three separate linear layers project the latent state, action, and task embeddings to a shared space.
    \item These three vectors are concatenated into a single vector.
    \item This vector is processed by an MLP with a final layer that outputs the scalar Q-value.
\end{enumerate}
The state-value function, $V_\psi(s)$, is trained separately but utilizes an identical architectural pattern, taking only the state and task embeddings as input. The modest size of these networks ($\sim$10-11M parameters) stands in stark contrast to the 3 billion parameters of the frozen VLM, making them significantly cheaper and faster to train and to query during inference. This efficiency is central to our method's practicality.
\subsection{Training objectives}
\label{sec:iql-training}

To train our Q-value model, we use an offline reinforcement learning approach based on Implicit Q-Learning (IQL) \cite{kostrikov_offline_2021}. This method is particularly well-suited for our setting because it can learn from a static, pre-collected dataset with sparse rewards and avoids the distributional shift problem inherent in many offline RL algorithms \cite{kostrikov_offline_2021}. IQL addresses this by training a value function that approximates the value of optimal, in-sample actions without ever explicitly querying out-of-sample actions \cite{kostrikov_offline_2021}. The training process involves alternating between two objectives. First, we learn a state-value function, $V_{\psi}(s)$, via expectile regression over the action-value function, $Q_{\theta}(s,a)$, using a loss that gives more weight to higher returns:
$$
L_{V}(\psi) = \mathbb{E}_{(s,a)\sim\mathcal{D}} \left[ L_{2}^{\tau}(Q_{\overline{\theta}}(s,a) - V_{\psi}(s)) \right],
$$

with 
$$
L_2^\tau(u) = |\tau - \mathbb{I}(u < 0)| u^2.
$$

A high value of the hyperparameter $\tau$ allows this objective to approximate the maximization operator over in-sample values, effectively performing multi-step dynamic programming \cite{kostrikov_offline_2021}. This is crucial for our sparse reward environment, where the model needs to "stitch" together parts of suboptimal trajectories to find a complete solution \cite{kostrikov_offline_2021}. Second, we use this learned value function to update the Q-function, minimizing a simple mean squared error (MSE) loss:

$$
L_{Q}(\theta) = \mathbb{E}_{(s,a,s^{\prime})\sim\mathcal{D}} \left[ \left(r(s,a) + \gamma V_{\psi}(s^{\prime}) - Q_{\theta}(s,a)\right)^2 \right].
$$

This approach allows the model to learn a near-optimal Q-function without the extrapolation errors that affect other methods. Once training is complete, the final policy is extracted using a separate advantage-weighted regression step \cite{kostrikov_offline_2021}.

In this framework, our decision to use Implicit Q-Learning (IQL) is a deliberate one, diverging from the standard Temporal Difference (TD) learning employed in methods like Digi-Q. The IQL algorithm is uniquely suited for offline reinforcement learning as it completely avoids querying the values of out-of-sample actions during training, thus circumventing the need for complex constraints or regularizations to manage distributional shift. This simplification makes our training process more stable and robust, as we do not need to explicitly account for the agent's behavior drifting from the data collection policy. Furthermore, IQL's use of expectile regression is particularly effective for our sparse reward environment. It enables the model to perform multi-step dynamic programming and learn an optimal value function by ``stitching'' together disparate parts of suboptimal trajectories, a task at which one-step methods often fail. This capability allows us to learn from a diverse dataset containing both successful and unsuccessful trajectories without the added complexity of a separate policy during the value learning phase. A breakdown of the hyperparameters used during training is given in Appendix \ref{sec:hyperparameters-training}.

\subsection{System Prompt of the Best-of-Q Agent}
\label{sec:multi-action-system-prompt}

The system prompt is engineered to elicit multiple, diverse action candidates from the base VLM. It guides the model toward strategic thinking by emphasizing three key aspects:
\begin{itemize}
    \item \textbf{Action Diversity:} The prompt explicitly requests multiple candidates and prioritizes the use of distinct \texttt{action} types when multiple strategic paths are viable.
    \item \textbf{Independent Reasoning:} Each candidate must be accompanied by its own \texttt{thought} and \texttt{note} fields to independently justify its strategic value.
    \item \textbf{Structured Output:} The prompt enforces a strict schema, ensuring the system can reliably parse the generated actions and their associated reasoning.
\end{itemize}
The complete system prompt used for this purpose is presented in Figure \ref{fig:multi_step_prompt}.

\begin{figure}[h!]
    \centering
    \begin{tcolorbox}[colback=gray!5, colframe=gray!75, title=\textbf{System Prompt}, fonttitle=\bfseries]
    \small
\begin{verbatim}
<multi_step_planning>
Generate {n_actions} **DIVERSE and STRATEGIC** complete step candidates.
**GUIDELINE: Prioritize proposing step candidates with different 'action' types to
explore varied strategies.**

- If multiple different action types are logical and strategic, you **MUST** use
them. For example, if both clicking a link and using a search bar are viable,
propose one of each.
- If only one 'action' type is truly sensible for the current state (e.g., the only
logical next step is to click one of several different buttons), you may propose
multiple candidates of the same action type. However, each candidate's 'thought'
**MUST** explain a different strategic goal or reason for choosing that specific
target.

Each step candidate should have:
- **Independent reasoning**: Each thought should justify why this specific approach
is valuable RIGHT NOW.
- **Contextual notes**: Each note should highlight info relevant to that particular
approach.
- **Diverse actions**: Choose from the available 'action' types to create varied
strategies whenever possible.

Available 'action' types:
- 'WriteElementRelativeAction'
- 'ClickElementRelativeAction'
- 'ScrollAction'
- 'GoBackWebAction'
- 'AnswerAction'
- 'WaitAction'
- 'RefreshWebAction'
- 'RestartAction'
</multi_step_planning>
\end{verbatim}
    \end{tcolorbox}
    \caption{System prompt used to generate diverse action candidates.}
    \label{fig:multi_step_prompt}
\end{figure}

\section{Detailed Experimental Setup}
\label{sec:hyperparameters-training}

This section provides a comprehensive overview of the experimental environment, hyperparameter configurations, and architectural details used to train the Q-function and V-function networks.

\subsection{Hyperparameters and Training Details}
The training of our lightweight Q-function and V-function was performed using a distributed setup on a single node equipped with 8 GPUs, utilizing the PyTorch \texttt{torchrun} utility. The training process spanned 50,000 epochs with a global batch size of 128. Central to our approach is Implicit Q-Learning (IQL), where the expectile parameter $\tau$ plays a critical role in balancing the estimation of the value function.

The full set of hyperparameters is detailed below:
\begin{itemize}
    \item \textbf{Optimizer}: Adam
    \item \textbf{Learning Rates}:
        \begin{itemize}
            \item Q-Network: $3 \times 10^{-4}$ with a cosine decay scheduler.
            \item V-Network: $3 \times 10^{-4}$ with a cosine decay scheduler.
        \end{itemize}
    \item \textbf{Expectile Parameter ($\tau$)}:
    \begin{itemize}
        \item For GPT-4.1 data: $\tau = 0.8$
        \item For Qwen2.5-VL data: $\tau = 0.7$ (Lowered to mitigate overestimation due to higher failure rates in the dataset).
    \end{itemize}
    \item \textbf{Target Network Update Frequency}: Every 100 gradient steps.
    \item \textbf{Discount Factor ($\gamma$)}: 0.99
    \item \textbf{Weight Decay}: $1 \times 10^{-4}$
    \item \textbf{Gradient Clipping}: Norm set to 1.0 to stabilize training.
\end{itemize}

\subsection{Model Architecture}
Both the Q-function and V-function are implemented as Multi-Layer Perceptrons (MLPs). Designed for computational efficiency, they operate on fixed-size embeddings extracted from the frozen VLM backbone. Each network contains approximately 11 million trainable parameters.

\paragraph{Q-Network Architecture.}
The Q-network accepts three inputs: a \textbf{state embedding}, an \textbf{action embedding}, and a \textbf{task embedding}. The forward pass proceeds as follows:
\begin{enumerate}
    \item \textbf{Projection:} Each of the three embeddings is projected into a shared latent dimension of 1024 using separate linear layers.
    \item \textbf{Concatenation:} The projected vectors are concatenated into a single joint representation.
    \item \textbf{MLP Processing:} The joint vector is passed through a feed-forward network with hidden layer dimensions of $[1024, 1024, 512, 256, 128, 64]$. ReLU activation and Dropout are applied after each hidden layer.
    \item \textbf{Output:} The final linear layer outputs a single scalar representing the $Q$-value.
\end{enumerate}

\paragraph{V-Network Architecture.}
The V-network estimates the state-value function $V(s)$ and follows a similar but simplified architecture, as it does not condition on the action embedding. It takes the \textbf{state} and \textbf{task} embeddings as input, projects them to the shared 1024 dimension, concatenates them, and processes the result through an MLP identical to the Q-network (same depth and width). The final layer outputs a single scalar $V$-value.

\section{WebVoyager Benchmark Details}
\label{sec:webvoyager-details}
The WebVoyager benchmark used in our experiments is based on the industry-standard set of 590 patched web-based tasks, designed to ensure tasks are robust against website changes on the live internet. These tasks are distributed across 15 distinct, real-world domains, each presenting unique challenges related to navigation, filtering, and content interaction. The breakdown of the total number of tasks evaluated per domain is detailed in Table \ref{tab:webvoyager_domain_breakdown}.

\begin{table}[h!]
    \centering
    \caption{Distribution of Tasks Across WebVoyager Domains (Total Tasks: 590)}
    \label{tab:webvoyager_domain_breakdown}
    \begin{tabular}{|l|c|}
        \hline
        \textbf{Domain} & \textbf{Total Tasks} \\
        \hline
        Allrecipes & 40 \\
        Amazon & 38 \\
        Apple & 34 \\
        ArXiv & 42 \\
        BBC-News & 34 \\
        Bing-Search & 40 \\
        Booking & 40 \\
        Cambridge-Dictionary & 43 \\
        Coursera & 40 \\
        ESPN & 41 \\
        GitHub & 39 \\
        Google-Flights & 39 \\
        Google-Map & 38 \\
        Huggingface & 36 \\
        Wolfram-Alpha & 46 \\
        \hline
    \end{tabular}
\end{table}

\section{Extended Results}
\label{sec:extended-results}
To complement the summary provided in the main report, this section offers a granular breakdown of the agent's success rate on a per-domain basis for the key models and methods evaluated. Results are averaged on 3 runs. We can see that for the three policies, the Best-of-Q outperformed almost all the domains. Exceptions are for Cambridge-Dictionary which was CAPTCHA-blocked when the agents were ran. Thus, no successful paths could be inferred from the training dataset and the Q-function was of no use on this domain for Qwen2.5-72B in particular.

\begin{table}[h!]
    \centering
    \caption{GPT-4.1 Agent Success Rates Per Domain on WebVoyager}
    \label{tab:gpt4_domain_sr}
    \begin{tabular}{|l|c|c|c|}
        \hline
        \textbf{Domain} & \textbf{Prompting} & \textbf{Random} & \textbf{Best-of-Q} \\
        \hline
        Allrecipes & 85.5\% & 69.0\% &\textbf{94.0\%} \\
        Amazon & 86.0\% & 74.6\% &\textbf{92.1\%} \\
        Apple & \textbf{95.8\%} & 78.6\% & 93.1\% \\
        ArXiv & 81.7\% & 75.8\% &\textbf{86.3\%} \\
        BBC-News & \textbf{90.2\%} & 68.1\% & 90.1\% \\
        Bing-Search & 83.1\% & 81.7\% &\textbf{89.1\%} \\
        Booking & 73.9\% & 44.4\% &\textbf{74.6\%} \\
        Cambridge-Dictionary & 63.4\% & 58.1\% &\textbf{79.4\%} \\
        Coursera & 91.7\% & 84.2\% &\textbf{96.7\%} \\
        ESPN & 73.0\% & 66.1\% &\textbf{85.8\%} \\
        GitHub & \textbf{94.9\%} & 82.7\% &89.7\% \\
        Google-Flights & 84.6\% & 51.7\% &\textbf{87.2\%} \\
        Google-Map &  92.1\% & 83.3\% &\textbf{92.9\%} \\
        Huggingface &  85.2\% & 83.2\% &\textbf{89.8\%} \\
        Wolfram-Alpha &  90.6\% & 92.8\% &\textbf{94.2\%} \\
        \hline
    \end{tabular}
\end{table}

\begin{table}[h!]
    \centering
    \caption{Qwen2.5-VL-7B Agent Success Rates Per Domain on WebVoyager}
    \label{tab:qwen7b_domain_sr}
    \begin{tabular}{|l|c|c|c|}
        \hline
        \textbf{Domain} & \textbf{Prompting} & \textbf{Random} & \textbf{Best-of-Q} \\
        \hline
        Allrecipes & 48.3\% & 52.5\% & \textbf{69.2\%} \\
        Amazon & 51.8\% & 59.5\% & \textbf{65.5\%} \\
        Apple & 52.9\% & 58.8\% & \textbf{63.6\%} \\
        ArXiv & 46.4\% & 52.4\% & \textbf{71.0\%} \\
        BBC-News & 36.6\% & 48.5\% & \textbf{51.0\%} \\
        Bing-Search & 30.5\% & 38.5\% & \textbf{65.2\%} \\
        Booking & 32.2\% & 27.8\% & \textbf{41.7\%} \\
        Cambridge-Dictionary & 15.9\% & 14.1\% & \textbf{20.8\%} \\
        Coursera & 55.0\% & 55.0\% & \textbf{71.2\%} \\
        ESPN & 20.0\% & 27.2\% & \textbf{40.0\%} \\
        GitHub & 50.9\% & 54.5\% & \textbf{64.1\%} \\
        Google-Flights & 25.9\% & 24.4\% & \textbf{42.7\%} \\
        Google-Map & 26.3\% & 31.6\% & \textbf{58.4\%} \\
        Huggingface & 41.7\% & 36.1\% & \textbf{56.5\%} \\
        Wolfram-Alpha & 46.4\% & 47.3\% & \textbf{60.3\%} \\
        \hline
    \end{tabular}
\end{table}

\begin{table}[h!]
    \centering
    \caption{Qwen2.5-VL-72B Agent Success Rates Per Domain on WebVoyager}
    \label{tab:qwen72b_domain_sr}
    \begin{tabular}{|l|c|c|c|}
        \hline
        \textbf{Domain} & \textbf{Prompting} & \textbf{Random} & \textbf{Best-of-Q} \\
        \hline
        Allrecipes & 68.1\% & 66.2\% & \textbf{84.6\%} \\
        Amazon & 85.1\% & 72.6\% & \textbf{87.6\%} \\
        Apple & 74.5\% & 68.2\% & \textbf{89.1\%} \\
        ArXiv & 68.0\% & 63.0\% & \textbf{79.3\%} \\
        BBC-News & 79.6\% & 74.6\% & \textbf{87.9\%} \\
        Bing-Search & 73.0\% & 60.3\% & \textbf{80.0\%} \\
        Booking & 58.8\% & 40.5\% & 58.8\% \\
        Cambridge-Dictionary & \textbf{36.1\%} & 27.2\% & 26.4\% \\
        Coursera & 78.3\% & 70.9\% & \textbf{85.8\%} \\
        ESPN & 51.2\% & 40.8\% & \textbf{67.8\%} \\
        GitHub & 77.8\% & 69.2\% & \textbf{89.7\%} \\
        Google-Flights & 71.6\% & 24.7\% & \textbf{77.8\%} \\
        Google-Map & 75.4\% & 68.4\% & \textbf{80.5\%} \\
        Huggingface & 67.0\% & 58.3\% & \textbf{71.3\%} \\
        Wolfram-Alpha & 79.7\% & 72.5\% & \textbf{88.4\%} \\
        \hline
    \end{tabular}
\end{table}

\clearpage

\section{Generalization}
In Reinforcement Learning, and in Deep Learning in general, one thing to consider is generalization capabilities of models. Here we can formulate generalization in two ways:
\begin{enumerate}
    \item Is the Best-of-Q Agent capable of generalizing to new tasks, and new domains (Section \ref{sec:gen-new-tasks})? 
    \item Is a Q-function trained on data collected with one model, able to generalize to the action space of another one (Section \ref{sec:gen-new-actions})?
\end{enumerate}

\subsection{Generalizing to new tasks and domains}
\label{sec:gen-new-tasks}
To test cross-domain generalization, we trained a Q-function on data collected from a set of web domains different from those in WebVoyager. To do this, we generated new tasks on new domains, using the original \textsc{Google-Flights} domain as a seed concept. The methodology for generating these new tasks is close to that of WebRL \cite{qi_webrl_2025}, where a few tasks from the original domain are used as seeds to prompt an LLM to generate new tasks on new domains.

This generalization dataset, that we call \textbf{Google-Flights Extended} is composed of three new domains, for which details can be seen in 
Table \ref{tab:gf-ext-details}. Particularly, Skyscanner and Momondo are flight booking websites, with different features and UIs than Google-Flights. Cruisecritic is a platform for booking cruises, which is a bit different from flights, but in essence, the tasks are close and its UI presents similar features. 

\begin{table}[h!]
    \centering
    \caption{Distribution of Tasks Across Google-Flights Extended (Total Tasks: 120)}
    \label{tab:gf-ext-details}
    \begin{tabular}{|l|c|}
        \hline
        \textbf{Domain} & \textbf{Total Tasks} \\
        \hline
        Skyscanner & 40 \\
        Momondo & 40 \\
        Cruisecritic & 40 \\
        \hline
    \end{tabular}
\end{table}

We followed the same data collection strategy as in Section \ref{sec:iterative-data-collection}, trained a Q-function, and benchmarked the Best-of-Q Agent on the \textsc{Google-Flights} domain of WebVoyager. For comparison, we also trained a Q-function on data collected directly from the original \textsc{Google-Flights} domain, with results reported in Table \ref{tab:generalization_results}. This experiment was conducted with both the Qwen2.5-VL-7B and Qwen2.5-VL-72B VLM policies.

\begin{figure}[h!]
    \centering
    \begin{subfigure}[b]{0.49\linewidth}
        \includegraphics[width=\linewidth]{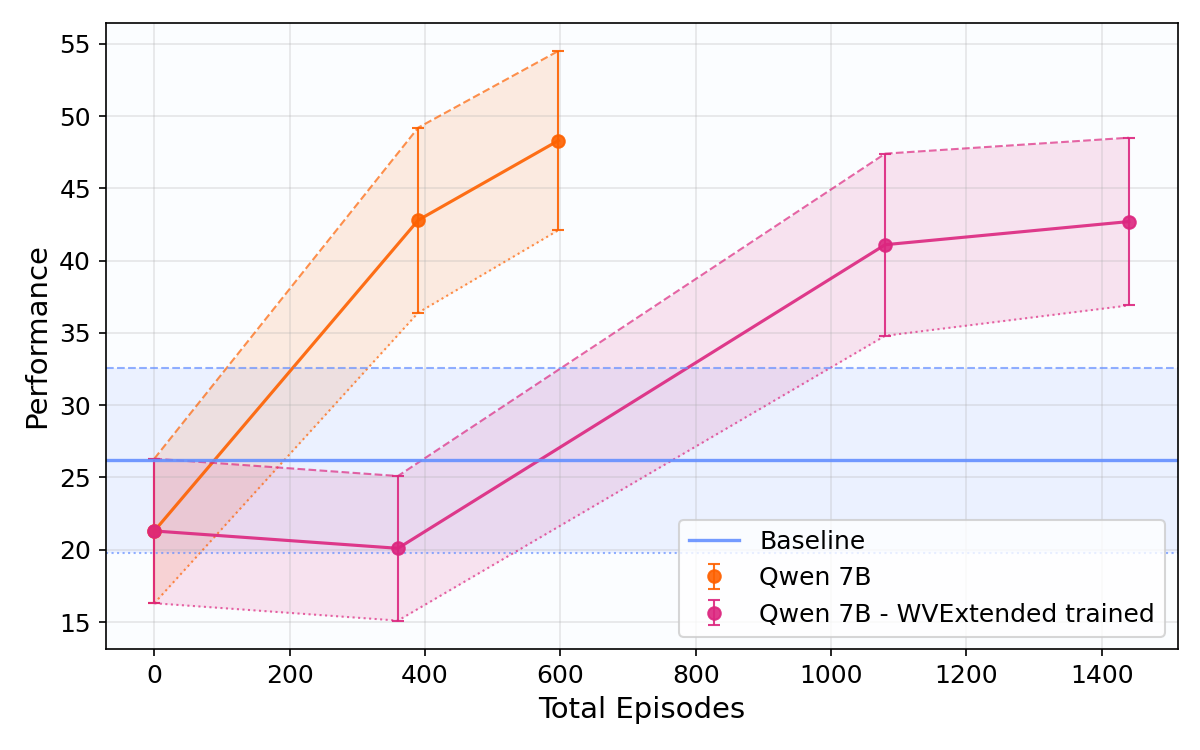}
        \caption{Evolution of the Best-of-Q Agent's performance on Google-Flights wrt total number of traces.}
        \label{fig:sample-efficiency-7b-generalization-all}
    \end{subfigure}
    \hfill
    \begin{subfigure}[b]{0.49\linewidth}
        \includegraphics[width=\linewidth]{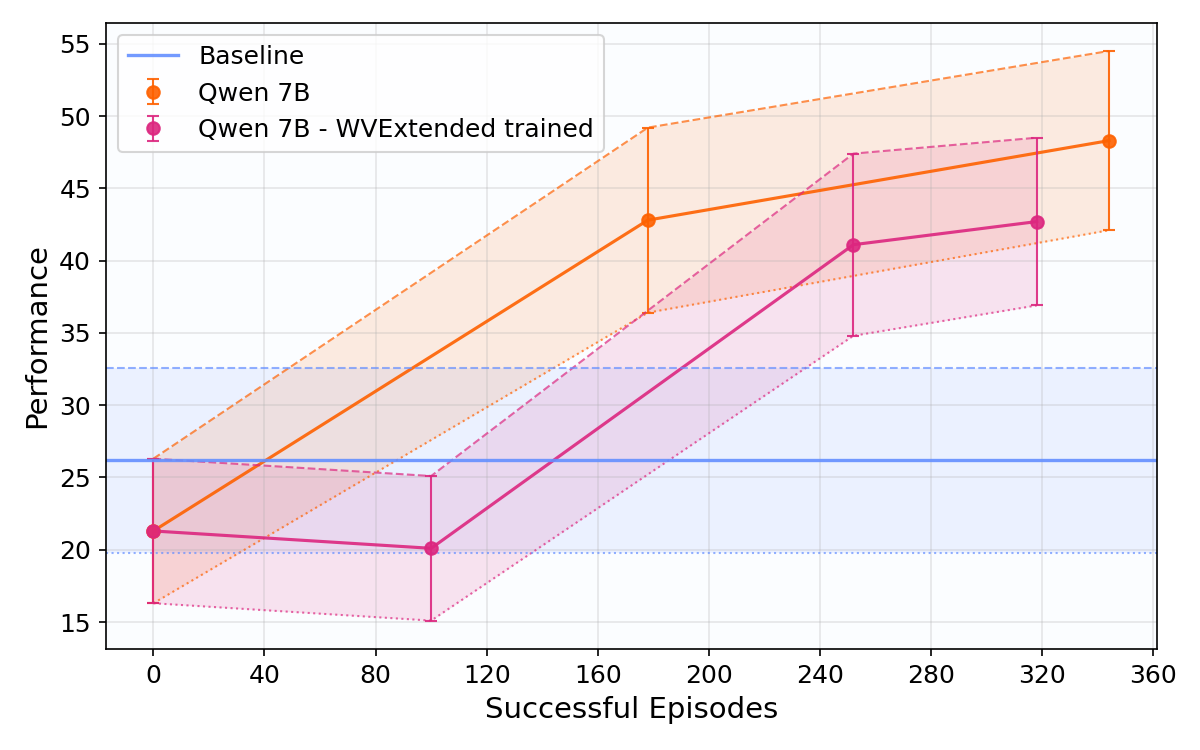}
        \caption{Evolution of the Best-of-Q Agent's performance on Google-Flights wrt number of successful traces}
        \label{fig:sample-efficiency-7b-generalization-success}
    \end{subfigure}
    \hfill
    \begin{subfigure}[b]{0.49\linewidth}
        \includegraphics[width=\linewidth]{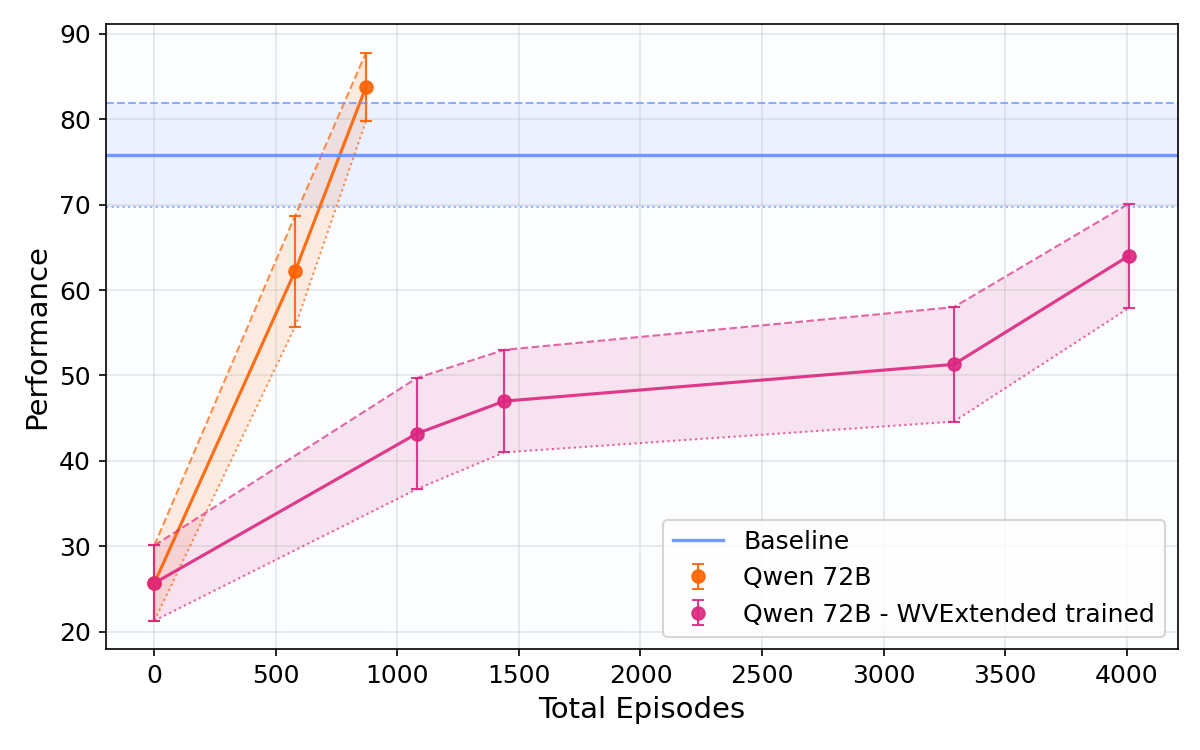}
        \caption{Evolution of the Best-of-Q Agent's performance on Google-Flights wrt total number of traces with Qwen2.5-vl-72B policy}
        \label{fig:sample-efficiency-72b-generalization-all}
    \end{subfigure}
    \hfill
    \begin{subfigure}[b]{0.49\linewidth}
        \includegraphics[width=\linewidth]{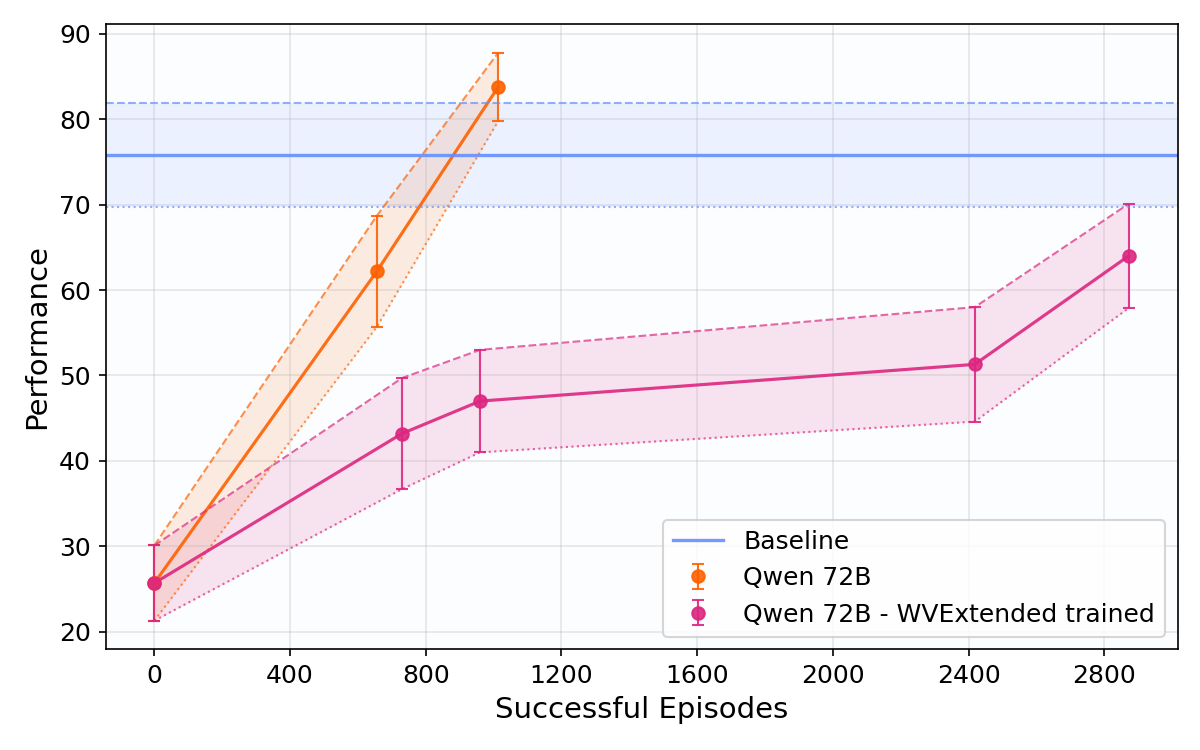}
        \caption{Evolution of the Best-of-Q Agent's performance on Google-Flights wrt number of successful traces with Qwen2.5-vl-72B policy.}
        \label{fig:sample-efficiency-72b-generalization-success}
    \end{subfigure}
    \caption{Sample efficiency of Best-of-Q Agent in generalization settings.}
    \label{fig:sample-efficiency-generalization}
\end{figure}

The results show that even when trained on a dataset of entirely out-of-domain tasks, the agent can transfer general knowledge of "booking", "browsing", or "comparing" flights, even across different UIs and website features. This comes with a downside: more data is needed to achieve a similar (or even slightly lower) performance. As shown in Figures \ref{fig:sample-efficiency-7b-generalization-all} and \ref{fig:sample-efficiency-72b-generalization-all}, a higher number of total training data points is needed to achieve comparable performance on \textsc{Google-Flights} when using out-of-domain data. We hypothesize the main reason is that the Google-Flights Extended tasks are harder than the original \textsc{Google-Flights} tasks, yielding fewer successful trajectories, which are key for training the Q-function. Indeed, if we compare performance based on the number of \textit{successful traces} in the training data (rather than total traces), this performance gap shrinks (Fig. \ref{fig:sample-efficiency-7b-generalization-success}). This suggests that curriculum learning methods could boost sample efficiency in such generalization experiments.

\begin{table*}[ht]
\centering
\begin{threeparttable}
\caption{Cross-Domain Generalization Results on flights booking tasks}
\label{tab:cross_domain_gen}
\begin{tabularx}{\textwidth}{|>{\raggedright\arraybackslash}X|l|c|}
\hline
\textbf{Method} & \textbf{Training Tasks} & \textbf{Success Rate ($\uparrow$)} \\
\hline
Qwen2.5-VL-7B Prompting & $-$ & 26.2 $\pm$ 6.4\% \\
\hline
Qwen2.5-VL-7B + Q-function & Google-Flights & 48.3 $\pm$ 6.2\% \\
\hline
Qwen2.5-VL-7B + Q-function & Google-Flights Extended  & 42.7 $\pm$ 5.8\% \\
% \hline
% Qwen2.5-VL-72B Prompting & $-$ & 75.8 $\pm$ 6.1\% \\
% \hline
% Qwen2.5-VL-72B + Q-function & Google-Flights & 83.8 $\pm$ 4.0\% \\
% \hline
% Qwen2.5-VL-72B + Q-function & Google-Flights Extended  & XX $\pm$ XX\% \\
\hline
\end{tabularx}
\end{threeparttable}
\end{table*}

\subsection{Generalization to another action space}
\label{sec:gen-new-actions}

Another interesting study is to try and use the Best-of-Q Agent with a policy (ex. Qwen2.5-VL-72B) and a Q-function that wasn't trained on data collected with that policy. One application to that is to see if we can reproduce a form of distillation from a bigger to a weaker model. We would collect data with a bigger model, and then use a weaker policy at inference, which would result in a cheaper and more compact agent.

\begin{table*}[h!]
\centering
\begin{threeparttable}
\caption{Generalization of Q-Function: Performance on Different Action Spaces}
\label{tab:generalization_results}
\begin{tabularx}{\textwidth}{|>{\raggedright\arraybackslash}X|l|l|c|c|}
\hline
\multicolumn{2}{|c|}{} & \multicolumn{2}{c|}{\textbf{Policy Model at Inference}} \\
\hline
\multicolumn{2}{|l|}{\textbf{Q-Function trained on data from \textbf{$\downarrow$} } } & \textbf{Qwen2.5-VL-72B} & \textbf{Qwen2.5-VL-7B} \\
\hline
\textbf{GPT4.1-trained} & Success Rate ($\uparrow$) & 69.5 $\pm$ 1.2\% & 43.1 $\pm$ 1.1\% \\
& Avg. Steps for Success ($\downarrow$) & 9.5 $\pm$ 0.3 & 10.2 $ \pm$ 0.3 \\
\hline
\textbf{Qwen2.5-VL-72B-trained} & Success Rate ($\uparrow$) & 76.6 $\pm$ 1.0\% & 45.6 $\pm$ 1.4\% \\
& Avg. Steps for Success ($\downarrow$) &  8.8 $\pm$ 0.2 & 10.4 $\pm$ 0.3 \\
\hline
\textbf{Qwen2.5-VL-7B-trained} & Success Rate ($\uparrow$) & 71.3 $\pm$ 1.1\% & 55.7 $\pm$ 1.3\% \\
& Avg. Steps for Success ($\downarrow$) & 9.6 $\pm$ 0.2 & 10.3 $\pm$ 0.3 \\
\hline
\end{tabularx}
\begin{tablenotes}
    \small
    \item \textbf{Success Rate ($\uparrow$)}: The percentage of tasks successfully completed. Higher is better.
    \item \textbf{Avg. Steps for Success ($\downarrow$)}: The average number of steps taken to complete successful tasks. Lower is better.
    \item \textbf{Note}: This table studies the performance of a Q-function trained on data from one policy when used to guide another policy at inference time.
\end{tablenotes}
\end{threeparttable}
\end{table*}

\label{sec:gf-extended-details}

\section{Reliability and Variance Analysis}
\label{sec:variance_analysis}

To understand \textit{why} our Best-of-Q method achieves a higher overall success rate than the standard prompting baseline (as shown in Table~\ref{tab:webvoyager_results}), we analyze the agent's reliability. VLM-based agents are inherently stochastic; running the standard \textbf{Prompting} agent on the same task multiple times often yields different outcomes, succeeding on some attempts and failing on others. This high variance in performance on a per-task basis suggests that the baseline's average success rate is suppressed by its unreliability.

Our hypothesis is that the Best-of-Q method improves the overall success rate not just by finding better actions, but by making the agent's decisions significantly more \textbf{reliable} and less stochastic.

\begin{figure}[!ht] 
    \centering
    \begin{subfigure}[b]{0.48\linewidth}
        \centering
        \includegraphics[width=\textwidth]{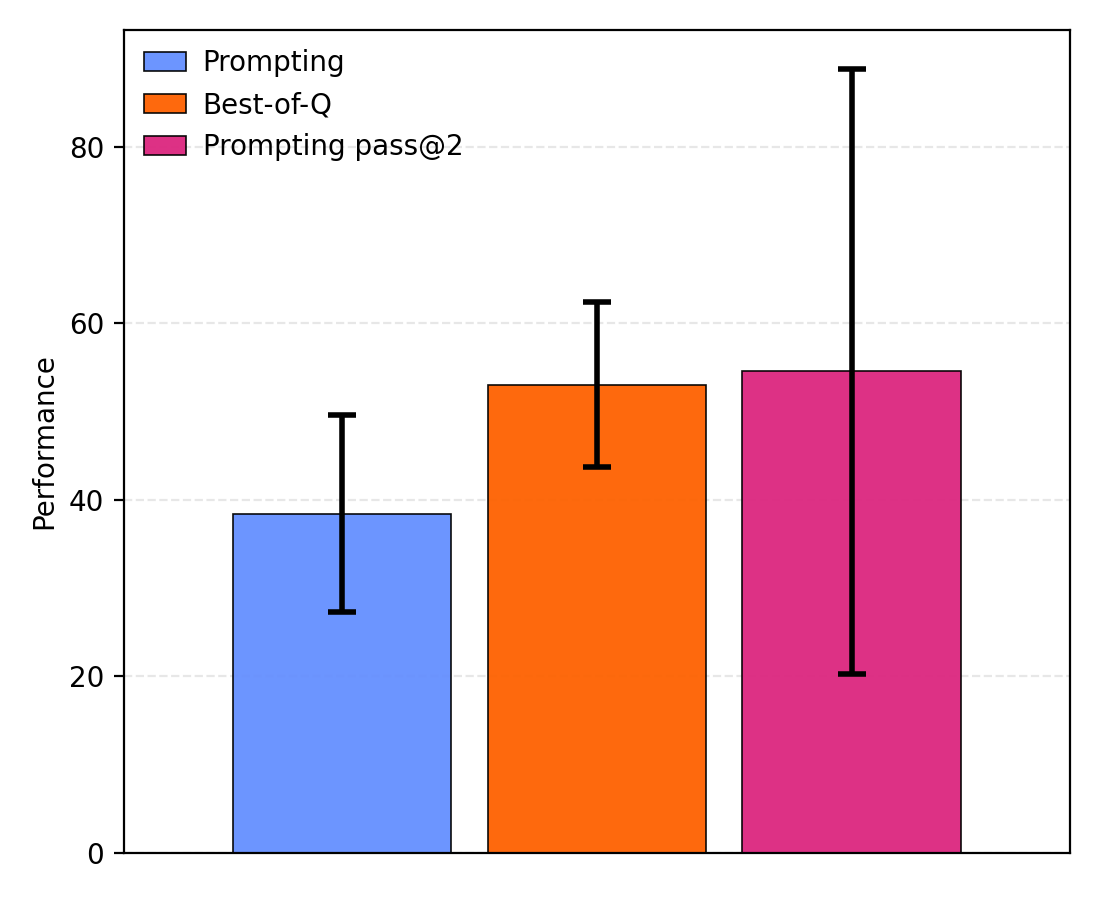}
        \caption{Qwen2.5-VL-7B}
        \label{fig:variance_7b}
    \end{subfigure}% <--- ADD THIS % TO REMOVE THE NEWLINE SPACE
    \hfill % Adds horizontal space
    \begin{subfigure}[b]{0.48\linewidth}
        \centering
        \includegraphics[width=\textwidth]{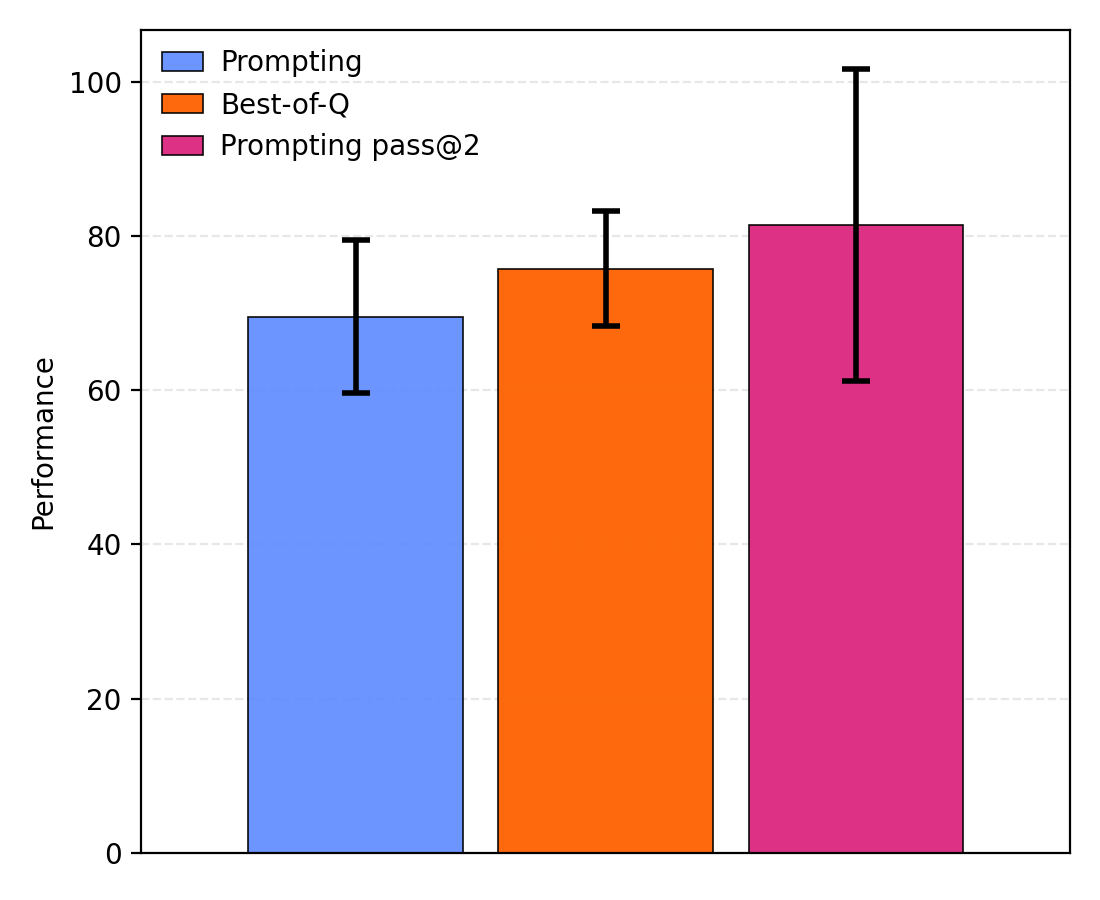}
        \caption{Qwen2.5-VL-72B}
        \label{fig:variance_72b}
    \end{subfigure}
    \caption{Aggregated task-level performance and variance comparison on all WebVoyager tasks. The error bars represent the 'mean of variances' across tasks, not the standard error of the benchmark. Our \textbf{Best-of-Q} method (pass@1) achieves performance comparable to a 'pass@2' baseline but with much lower task level variance, indicating higher reliability.}
    \label{fig:variance_compare}
\end{figure}

A common strategy to mitigate this baseline stochasticity is using a "pass@k" approach, where the agent is given $k$ retries. As seen in Figure~\ref{fig:variance_compare}, the \textbf{Prompting pass@2} baseline (which allows two retries) achieves a mean performance comparable to our \textbf{Best-of-Q} method. However, this comes at the cost of extremely high variance, indicating that its success is unreliable and highly dependent on "getting lucky" in one of its attempts.

Figure~\ref{fig:variance_compare} plots the aggregated task-level variance for the 7B and 72B models. It is important to note that the variance (error bars) in these charts represents an \textbf{aggregated variance across tasks} (a "mean of variances" for individual tasks), not the standard error of the entire benchmark, which is reported in Table \ref{tab:webvoyager_results} for example. The data reveals two key insights:
\begin{itemize}
    \item \textbf{Reduced Variance vs. pass@1:} Our \textbf{Best-of-Q} agent (green bar) consistently shows lower variance than the standard \textbf{Prompting} (pass@1) baseline.
    \item \textbf{Superior Reliability vs. pass@2:} More critically, our method achieves performance on par with the \textbf{Prompting pass@2} baseline but with \textbf{drastically lower variance}.
\end{itemize}

This analysis demonstrates that our Best-of-Q method makes the agent's performance more consistent. It effectively \textbf{de-randomizes the suboptimal stochasticity} of the base VLM. Instead of relying on multiple, high-variance retries to find a good trajectory, our method uses the Q-function to more deterministically select the best available action in a \textbf{single pass}. It delivers the high success rate of a multi-try strategy but with the reliability and low variance of a single, robust agent.

\section{Ablations}

Two ablations can be found in the main paper in Section \ref{sec:ablations}, below we provide an additional study on the influence of the embedder backbone.

\paragraph{\textbf{Influence of the Embedder Backbone}}
\label{sec:influence-embedder}

A core component of the Q-function model is its embedder. For our main results, we used the Qwen2.5-VL-3B VLM as the embedder due to its reasonable size and low inference cost. A natural question arises: would increasing the size and quality of this VLM embedder yield better results by providing a more meaningful feature space for the Q-function?

To investigate this, we trained a new Q-function on the same data but used a larger Qwen2.5-VL-7B model as the feature extractor. We then compared the performance of the Best-of-Q Agent using both the Qwen2.5-VL-3B and Qwen2.5-VL-7B embedders. The agent's policy backbone for both experiments remained the same ( Qwen2.5-VL-7B).

Table \ref{tab:influence-size-embedder} shows the results of this experiment. We observe that increasing the size of the embedder does not lead to a significant changes in performance. The success rate stays roughly the same (within standard deviation) with a slight decrease from 55.7\% to 55.0\%, and the average steps for success remain comparable. This finding supports our hypothesis from the failure analysis in Section \ref{sec:failure_modes}, which suggests that the agent's performance is fundamentally bounded by the base VLM's capacity to propose a sufficient set of high-quality action candidates, not by the Q-function's ability to select the best one from that set.

\begin{table*}[ht]
\centering
\caption{Comparison of Best-of-Q Agent Performance with Different Embedder Backbones}
\label{tab:influence-size-embedder}
\begin{tabularx}{\textwidth}{|>{\raggedright\arraybackslash}X|c|c|c|}
\hline
\textbf{Q-function Embedder} & \textbf{Embeddings size} & \textbf{Success Rate ($\uparrow$)} & \textbf{Avg. Steps for Success ($\downarrow$)} \\
\hline
Qwen2.5-VL-3B & 2048 & 55.7 $\pm$ 1.4\% & 9.7 $\pm$ 0.3 \\
Qwen2.5-VL-7B & 3584 & 55.0 $\pm$ 1.3\% & 9.5 $\pm$ 0.3 \\
Holo1-7B & 3584 & 55.3 $\pm$ 1.4\% & 9.6 $\pm$ 0.3 \\
\hline
\end{tabularx}
\end{table*}

\section{Prompt for Quantitative Failure Analysis}
\label{sec:quant-study-prompt}

In Section \ref{sec:failure_modes}, we conducted a quantitative analysis using GPT-4.1 to determine how frequently a "golden" action (derived from an expert trajectory) appeared within the candidate set proposed by the Qwen2.5-VL-7B policy. This analysis focused specifically on the Google Flights domain. Figure \ref{fig:quant-study-prompt} presents the exact system prompt used to perform this semantic matching evaluation.

\begin{figure}[h!]
    \centering
    \begin{tcolorbox}[colback=gray!5, colframe=gray!75, arc=2mm, boxrule=1pt]
    \small \ttfamily
    You are checking whether a FULL STEP (action + note + thought)\\
    is semantically present in a list of STEP CANDIDATES.\\
    
    Consider action type and key fields (selector/text/url/x,y),\\
    and also whether note and thought express the same intent/semantics.\\
    Ignore superficial formatting differences.\\
    
    STEP:\{vlm\_action\}\\
    CANDIDATES (ordered):\{action\_candidates\}\\
    
    Return JSON with fields: contained (bool),\\
    matched\_index (0-based or null), rationale (short).
    \end{tcolorbox}
    \caption{System prompt used to verify if the "golden" action is present in the VLM's candidate list.}
    \label{fig:quant-study-prompt}
\end{figure}

\section{Prompt for VLM-based Action Selection}
\label{sec:vlm-selector-prompt}

In Section \ref{sec:vlm-as-sampler}, we presented an analysis where a VLM is utilized as a zero-shot action selector, serving as a baseline to our trained Q-function. Figure \ref{fig:vlm-selector-prompt} provides the specific system prompt used to instruct the VLM to evaluate and rank the generated action candidates during inference.

\begin{figure}[h!]
    \centering
    \begin{tcolorbox}[colback=gray!5, colframe=gray!75, arc=2mm, boxrule=1pt]
    \small \ttfamily
    You are an expert in action ranking for web agent tasks.\\
    Read the full conversation and available actions below,\\
    then select the BEST option number.\\
    
    Step \{step\_index + 1\} of the episode.\\
    
    \{actions\_text\}\\
    
    Respond using the specified JSON schema.
    \end{tcolorbox}
    \caption{System prompt utilized for the VLM-based action selector baseline.}
    \label{fig:vlm-selector-prompt}
\end{figure}

\section{Cost analysis details}
\label{sec:cost-details}

We provide the values used to compute the cost of a benchmark run with different policies. The table \ref{tab:vlm_token_costs} shows what values we used to compute the cost of a benchmark for each VLMs. To have the final number, we counted the number of input tokens and output tokens needed to perform all the tasks (failures and successes) and averaged the results on 3 runs.
\begin{table*}[h!]
\centering
\caption{Estimated Token Costs for VLM Models (per 1M tokens)}
\label{tab:vlm_token_costs}
\begin{tabular}{|l|c|c|}
\hline
\textbf{VLM Model} & \textbf{Input Token Cost} & \textbf{Output Token Cost} \\
\hline
GPT-4.1 & \$2.00 & \$8.00 \\
\hline
Qwen2.5-VL-72B & \$1.00 & \$4.00 \\
\hline
Qwen2.5-VL-7B & \$0.15 & \$0.60 \\
\hline
Qwen2.5-VL-3B & \$0.10 & \$0.40 \\
\hline
\end{tabular}
\end{table*}

\section{Qualitative Analysis of Agent Trajectories}
\label{sec:qualitative-study}

To provide a clear qualitative understanding of the performance disparity and the fundamental bottleneck identified in our failure analysis (Section 4.3), we visualize the step-by-step value evolution of a successful agent (GPT-4.1) versus a failed agent (Qwen2.5-VL-7B) on the challenging Google Flights task: \textit{"Find flights from Chicago to London on 20 December and return on 23 December."}

In both figures below, the \textbf{V-Value} (dashed blue line) represents the predicted maximum future return from the current state, and the \textbf{Selected Action Path} (solid red line) represents the predicted value of the specific action chosen by the Best-of-Q reranker. The small dots represent the Q-values of the other candidate actions proposed by the VLM at that step.

\begin{figure}[h!]
    \centering
    \begin{subfigure}[b]{0.75\linewidth}
        \includegraphics[width=\linewidth]{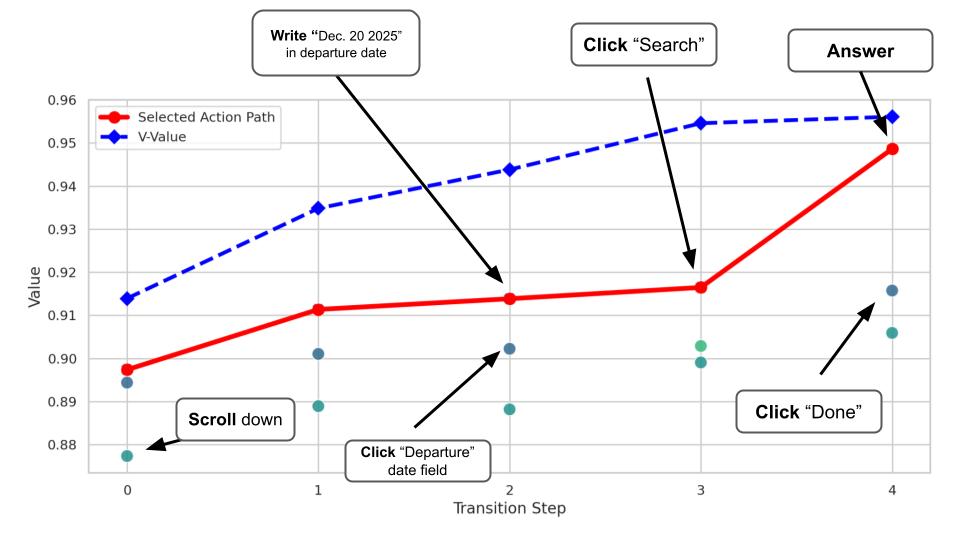}
        \caption{Successful Trajectory with GPT-4.1 Agent (Golden Standard). The VLM is highly efficient at successfully completing the task.}
        \label{fig:gpt4_trajectory}
    \end{subfigure}
    \vspace{0.5cm}
    \begin{subfigure}[b]{0.75\linewidth}
        \centering
        \includegraphics[width=\linewidth]{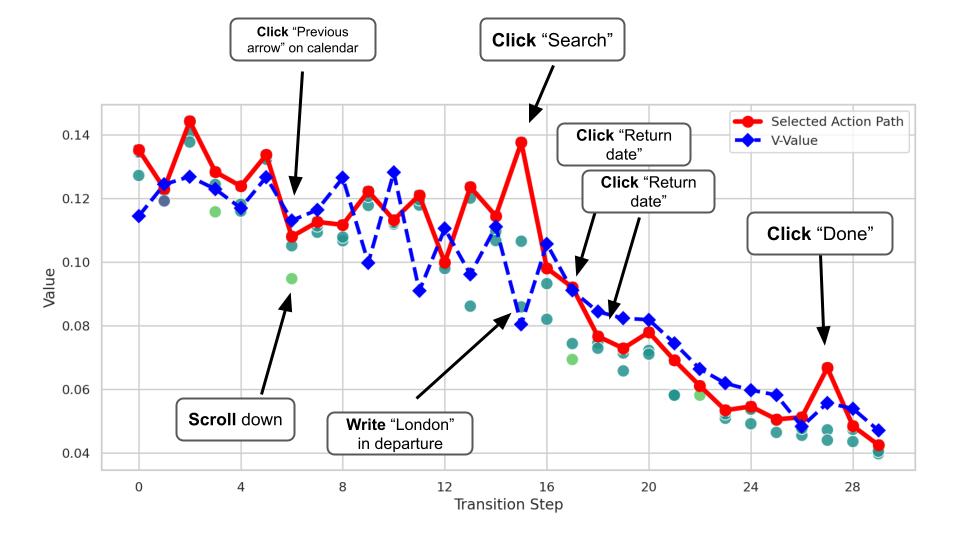}
        \caption{Failed Trajectory with Qwen2.5-VL-7B Agent.}
        \label{fig:qwen7b_trajectory}
    \end{subfigure}
    \caption{Comparison of trajectories between a GPT-4.1 (successful) and Qwen2.5-VL-7B (failed) policy.}
\end{figure}

\paragraph{Golden Standard Trajectory: GPT-4.1.}
Figure \ref{fig:gpt4_trajectory} illustrates a successful, highly efficient trajectory guided by the GPT-4.1 VLM. This trajectory is characterized by:
\begin{itemize}
    \item \textbf{High and Stable Values}: Both the V-Value (state quality) and the Selected Action Q-Value remain high (above 0.9) throughout the task. This indicates that the base GPT-4.1 VLM is consistently proposing high-quality, optimal actions.
    \item \textbf{Effective Reranking}: The Q-function effectively identifies and selects the best action at each step, maintaining the agent's progress toward the sparse reward (Success at $t=4$).
    \item \textbf{Efficiency}: The task is completed in just 4 steps, demonstrating robust, targeted planning where the VLM immediately proposes relevant actions like writing the date, clicking the search button, and answering the prompt.
\end{itemize}

\paragraph{Failed Trajectory: Qwen2.5-VL-7B.}
Figure \ref{fig:qwen7b_trajectory} illustrates a representative failed trajectory guided by the Qwen2.5-VL-7B VLM. In stark contrast to the GPT-4.1 agent, this agent suffers from the "VLM bottleneck":
\begin{itemize}
    \item \textbf{Value Decay and Instability}: The values start relatively low (around 0.12--0.14) and exhibit immediate, dramatic drops, decaying steadily toward zero. This decay is driven by the base VLM repeatedly offering low-quality, suboptimal, or repetitive actions.
    \item \textbf{Q-Function Constraint}: The Q-function operates correctly by assigning low values to all available options and selecting the least-bad choice from a suboptimal set. However, it cannot generate an effective action. For instance, the agent gets stuck in loops (e.g., repeatedly clicking the "Previous arrow" or clicking the "Return date" field multiple times without making progress).
    \item \textbf{Absence of Goal-Oriented Actions}: Crucially, the VLM fails to propose the necessary high-value actions required to complete the task within the initial steps, forcing the agent down a lengthy (30 steps) and ultimately unsuccessful path. The Q-function's impact is fundamentally capped by the poor quality of the raw action proposals from the base VLM.
\end{itemize}

The visual data strongly reinforces our central finding: the primary limitation on agent performance is the ability of the frozen VLM to propose a viable, high-quality action in the first place, rather than the reranking capability of our trained Q-function.

\end{document}